\documentclass{article}

\usepackage[preprint]{neurips_2026}

\usepackage[utf8]{inputenc}
\usepackage[T1]{fontenc}
\usepackage{url}
\usepackage{booktabs}
\usepackage{nicefrac}
\usepackage{microtype}
\usepackage{graphicx}
\usepackage{subfig}
\usepackage{wrapfig}
\usepackage{multirow}
\usepackage{amsmath, amssymb, amsfonts, amsthm}
\usepackage[table]{xcolor}
\usepackage{adjustbox}
\usepackage{makecell}
\usepackage{algorithm}
\usepackage{algorithmic}
\definecolor{MyGreen}{HTML}{228B22}
\usepackage{mmstyles}
\newcommand{\BlueComment}[1]{\hfill \textcolor{blue}{$\triangleright$~#1}}
\setcitestyle{round}

\definecolor{mydarkblue}{rgb}{0,0.08,0.45}

\setcitestyle{authoryear,round,citesep={;},aysep={,},yysep={;}}

\makeatletter
\renewcommand{\NAT@open}{\begingroup\color{mydarkblue}(}
\renewcommand{\NAT@close}{)\endgroup}
\makeatother

\usepackage{hyperref}
\title{Bridging the Semantic-Action Gap in Visual Token Pruning for Efficient VLA Inference}

\author{%
  \textbf{Ziyan Liu\textsuperscript{1,2}\thanks{Equal contribution.}}
  \quad
  \textbf{Yeqiu Chen\textsuperscript{1,2}\footnotemark[1]}
  \quad
  \textbf{Yiming Zhang\textsuperscript{1}}
  \quad
  \textbf{Hongyi Cai\textsuperscript{1}}
  \quad
  \textbf{Tao Lin\textsuperscript{1}}
  \\[0.4em]
  \textbf{Runquan Gui\textsuperscript{2}}
  \quad
  \textbf{Shuo Yang\textsuperscript{3}}
  \quad
  \textbf{Zheng Liu\textsuperscript{4}}
  \quad
  \textbf{Bo Zhao\textsuperscript{1}\thanks{Corresponding author: \texttt{bo.zhao@sjtu.edu.cn}.}}
  \\[0.8em]
  \normalfont
  \textsuperscript{1}School of AI, Shanghai Jiao Tong University
  \\
  \textsuperscript{2}University of Science and Technology of China
  \\
  \textsuperscript{3}Harbin Institute of Technology (Shenzhen)
  \\
  \textsuperscript{4}BAAI
}

\begin{document}

\maketitle

\begin{abstract}
Vision-Language-Action (VLA) models have shown great potential for embodied AI by integrating visual perception, language understanding, and action execution.
In real-time deployment, these models must process continuous visual streams, incurring substantial computational overhead.
Visual token pruning---a mainstream technique for accelerating Vision-Language Models (VLMs) by retaining salient tokens while discarding redundant ones---offers a natural candidate solution to this challenge. 
However, directly applying VLM-oriented pruning methods to VLA inference can cause severe degradation in manipulation performance. 
Our analysis attributes this degradation to a key mismatch: VLA inference exhibits distinct attention patterns between the vision-language prefill stage and the action-decode stage, so pruning based only on context-prefill \textit{semantic salience} is biased toward semantic cues and may remove action-critical visual tokens. 
Motivated by this observation, we propose \textbf{VLA-Pruner}, an effective plug-and-play token pruning method grounded in the visual requirements of VLA inference, further exploiting the temporal continuity of robot manipulation. 
Specifically, VLA-Pruner estimates visual-token importance from both semantic prefilling and temporally smoothed action relevance, and then applies a Combine-then-Filter strategy to retain compact, non-redundant tokens under the compute budget. 
Experiments show that VLA-Pruner outperforms state-of-the-art approaches across multiple VLA architectures, achieving up to 1.99$\times$ speedup with comparable manipulation quality. We release our code at \textcolor{mydarkblue}{\url{https://github.com/MINT-SJTU/VLA-Pruner}}.
\end{abstract}

\section{Introduction}

Recent advances in robot learning have led to remarkable progress in developing generalist policies capable of acting across diverse tasks and environments~\citep{ma2024survey,firoozi2025foundationsmodel_survey,zhao2025cot,brohan2022rt,zitkovich2023rt,kim2024openvla,black2024pi_0,li2024cogact,chi2025diffusionpolicy}. Vision-Language-Action (VLA) models~\citep{brohan2022rt,zitkovich2023rt,kim2024openvla,kim2025fine,black2024pi_0,li2024cogact,lin2025evo1}, which benefit from the multimodal understanding and transfer ability of Vision-Language Models (VLMs)~\citep{liu2023visulinstructiontuning,bai2023qwen}, have emerged as a promising pathway toward general-purpose robot policy. Leveraging large-scale real-world robotic datasets~\citep{o2024openx-embodiment}, VLA models can perceive visual scenes, understand language instructions, and execute low-level actions across diverse scenarios. However, real-time deployment requires VLA models to process continuous visual streams in dynamic environments, incurring heavy computational overhead. Existing VLA acceleration techniques, such as model lightweighting~\citep{wen2025tinyvla}, quantization~\citep{park2024quantizationvla}, and early-exit~\citep{yue2024deer}, mitigate this cost but often require architectural modifications or retraining, limiting their generalizability. 

A particularly appealing direction for training-free VLA acceleration is to reduce the computation induced by visual tokens. Prior VLM studies have shown that visual tokens are highly redundant~\citep{chen2024fastv,shang2024prumerge,alvar2025divprune},  and visual token pruning, which retains the most salient tokens while pruning the rest, has become a general and effective approach to improve VLM inference efficiency~\citep{alvar2025divprune,chen2024fastv,zhang2024sparsevlm,shang2024prumerge,ye2025fitprune,xing2024drop,holov}. Since modern VLA models typically build on VLM backbones and inherit their visual-token processing pipeline, visual token pruning shows promise for VLA acceleration. Existing visual token pruning methods typically define \textbf{importance criteria} for tokens, such as attention scores~\citep{chen2024fastv,zhang2024sparsevlm,holov}, gradient information~\citep{mao2025pruneandmerge,mao2025efficientvlm} or feature diversity~\citep{alvar2025divprune} to quantify the significance of visual tokens, and less important tokens are pruned at inference time. By reducing visual redundancy, they yield substantial compute savings with minimal accuracy loss on standard VLM tasks such as image captioning~\citep{agrawal2019nocaps,plummer2015flickr30k}, VQA~\citep{hudson2019gqa,liu2024mmbench}, and video understanding~\citep{xu2017video,li2016tgif}.

\begin{wrapfigure}{r}{0.48\textwidth}
    \vspace{-1em}
    \centering
    \includegraphics[width=\linewidth]{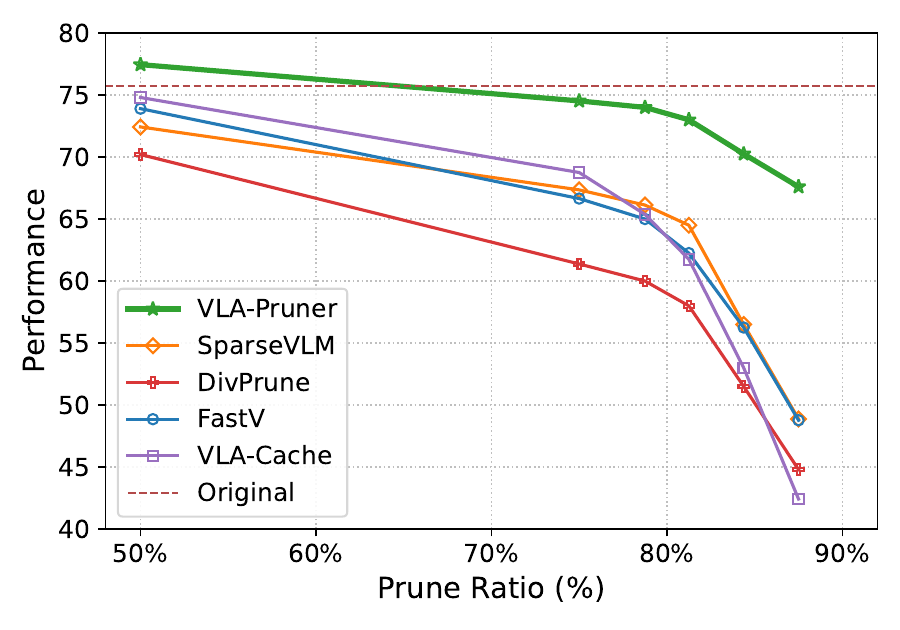}
    \vspace{-2em}
    \caption{\small Comparison of visual token pruning/caching methods across various pruning ratios. Directly applying VLM pruning methods to VLA inference causes severe performance degradation.}
    \label{fig:general-comparison}
    \vspace{-1em}
\end{wrapfigure}

Despite recent progress, directly applying VLM-oriented visual token pruning methods to VLA inference can lead to severe degradation in manipulation performance, especially at high pruning ratios (see Figure~\ref{fig:general-comparison}), as also observed in prior work~\citep{xu2025vlacache}. We analyze the source of this degradation and identify a key mismatch: VLA inference involves both high-level semantic understanding and task planning, as well as low-level action execution, which induce distinct visual requirements across the \textbf{vision-language prefill} and \textbf{action decode} stages. Correspondingly, we observe markedly different attention patterns between these two stages: prefill attention tends to reflect semantic relevance in the vision-language context, whereas action-decode attention captures visual information needed for precise action generation. However, existing VLM-oriented pruning methods typically rank visual tokens using \textit{semantic salience} measured during context prefilling---e.g., prefill attention in FastV~\citep{chen2024fastv} or text-to-vision cross-attention scores in SparseVLM~\citep{zhang2024sparsevlm}. Such semantic-only criteria bias token retention toward tokens salient for prefilling, thereby pruning action-critical visual tokens and limiting their applicability to VLA models.


To bridge this gap, we propose \textbf{VLA-Pruner}, a plug-and-play token pruning approach that accounts for stage-wise visual requirements andexploits the temporal continuity of robot manipulation. First, it performs \textbf{semantic-action importance estimation} for visual token retention: vision-language prefill attention scores measure semantic relevance, following prior VLM pruning works~\citep{chen2024fastv,ye2025fitprune,mao2025pruneandmerge,zhang2024sparsevlm}, while action-to-vision attention scores measure action relevance. Since current action-to-vision attention is unavailable when pruning is performed during prefilling, VLA-Pruner estimates it from recent decode attentions using a temporal smoother, motivated by the strong short-term overlap of top-attended visual patches across consecutive control steps. Second, VLA-Pruner adopts a \textbf{Combine-then-Filter token selection strategy} to avoid fragile score fusion. It first takes the union of tokens salient for either semantic prefilling or action decoding to maximize relevance, and then filters redundant candidates to obtain a compact token set under the given compute budget. This design reduces computational overhead while preserving visual information needed for both semantic understanding and action execution. 

We summarize our contributions. (1) We systematically analyze visual-token importance in VLA inference, revealing distinct attention patterns between semantic prefilling and action decoding as well as temporal consistency in action-to-vision attention, thereby identifying the limitations of current VLM pruning methods for VLA acceleration. (2) We propose VLA-Pruner, a general, training-free framework that combines semantic-action importance estimation with a Combine-then-Filter token selection strategy, exploiting temporal continuity to preserve action-relevant visual information. (3) As a plug-and-play module for multiple VLA models, VLA-Pruner achieves state-of-the-art performance across two simulated environments, delivering up to 1.99$\times$ speedup with negligible performance drops and remaining robust even at an 87.5\% pruning ratio. We also validate its real-world advantage on a 6-DoF xArm6 robot.

\section{Preliminary}
\paragraph{VLA Inference.} Vision-Language-Action (VLA) models process sequential inputs of the form $\{ V^{t}, L^{{t}} \}$ into $A^t$ across timesteps $t$. $V^{t}$ represents the input visual observations, $L^{t}$ denotes language task instructions, and $A^t = \{a^t_{1},...,a^t_{\hat{N}}\}$ denotes the predicted action tokens, which are detokenized into executable robotic actions. The text input is tokenized into $N$ text tokens $\mathbf{E}^{t}_\tau = \{\tau^t_1, \ldots, \tau^t_N\}$. The vision encoder processes $V^{t}$ into image features, which are projected into $M$ visual tokens $\mathbf{E}^t_v = \{v_1, \ldots, v_M\}$. These tokens, along with a proprioceptive token representing robot state, form the Transformer input. Since self-attention scales quadratically with token counts, the $M$ (generally $M \gg N$) visual tokens dominate inference cost.

\paragraph{VLA Attention Breakdown.} Like VLMs, VLA models rely on the attention mechanism of transformer for token interactions. Without loss of generality, we describe the single-head attention below. At each layer $l$, the model takes previous-layer hidden states of visual tokens
$\mathbf{H}^{l-1}_v \in \mathbb{R}^{M \times d}$ 
and textual tokens
$\mathbf{H}^{l-1}_\tau \in \mathbb{R}^{N \times d}$ 
as inputs. \textit{For mathematical simplicity, we omit time/layer superscripts and treat the proprioceptive token as part of the vision-language context}. These inputs are transformed into queries $\mathbf{Q}$, keys $\mathbf{K}$, and values $\mathbf{V}$ using linear projections. Then, the shared vision-language context for attention and interaction is formed as:  $\mathbf{K}_{\text{vl}} \in \mathbb{R}^{(N+M) \times d_k}, 
\mathbf{V}_{\text{vl}} \in \mathbb{R}^{(N+M) \times d_v}$. We break down VLA inference into two stages:\\
\noindent\textbf{(1) Vision-Language prefill Stage.}  
In this stage, queries come from the text and vision tokens:
$\mathbf{Q}_{\text{vl}} \in \mathbb{R}^{(N+M) \times d_k}$.  
The attention matrix is computed by
\begin{equation}
\abovedisplayskip=6pt minus 3pt
    \mathcal{A}_{\text{vl}} 
    = \text{Softmax}\!\left(
    \frac{\mathbf{Q}_{\text{vl}} \mathbf{K}_{\text{vl}}^{\top}}
    {\sqrt{d_k}}
    \right).
\belowdisplayskip=6pt minus 3pt
\end{equation}
Attention score for each input visual patch is computed as
\begin{equation}
\abovedisplayskip=6pt minus 4pt
    \mathcal{S}_{\text{vl}}[m] 
    = \frac{1}{M+N}\sum^{M+N}_{i=1} \mathcal{A}_{\text{vl}}[i, m], 
   \text{ }m=1,\dots,M.
\belowdisplayskip=6pt minus 4pt
\label{eq:prefill attention}
\end{equation}
\noindent\textbf{(2) Action decode Stage.}  
In this stage, queries are action tokens $\mathbf{Q}_{\text{act}} \in \mathbb{R}^{K \times d_k}$ ($K=1$ for autoregressive~\citep{kim2024openvla}, $K=\hat{N}$ for chunk-based~\citep{black2024pi_0}). Attention is computed as
\begin{equation}
\abovedisplayskip=6pt minus 4pt
    \mathcal{A}_{\text{act}} 
    = \text{Softmax}\!\left(
    \frac{\mathbf{Q}_{\text{act}} \mathbf{K}_{\text{vl}}^{\top}}
    {\sqrt{d_k}}
    \right).
\belowdisplayskip=6pt minus 4pt
\label{eq:action attention}
\end{equation}
For the autoregressive case, attention vectors from $\hat{N}$ decoding steps are concatenated into the final attention matrix. The action-to-vision attention score in this stage is
\begin{equation}
    \mathcal{S}_{\text{act}}[m] 
    = \frac{1}{\hat{N}} \sum_{i=1}^{\hat{N}} \mathcal{A}_{\text{act}}[i, m], 
    \text{ } m=1,\dots,M.
\end{equation}

\section{Analyzing Visual Token Importance in VLA Inference}
\label{sec:observation}

\begin{figure*}[t]
    \centering
    \subfloat[\small Average Overlap\label{fig:overlap_ratio}]{%
        \includegraphics[width=0.29\textwidth]{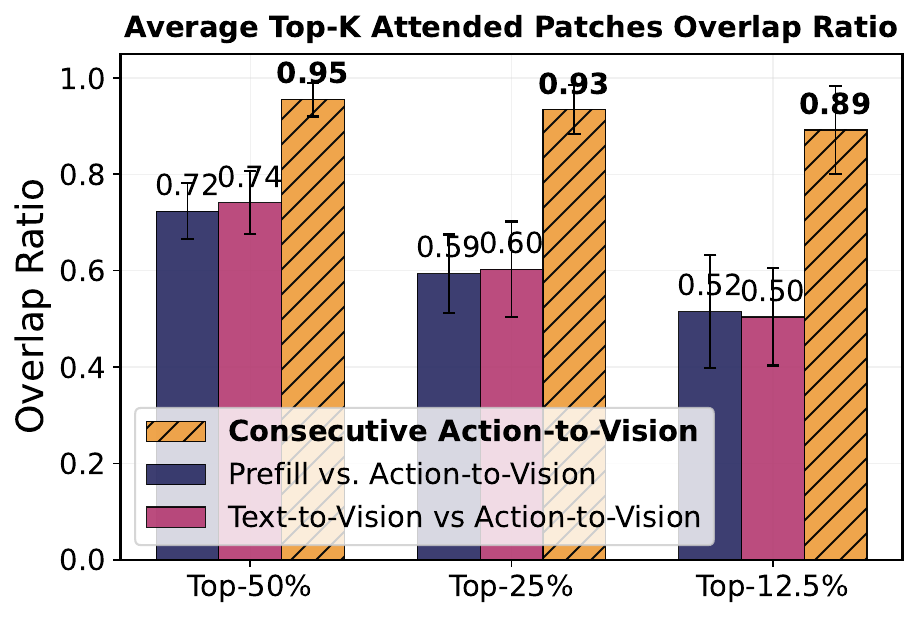}%
        \vspace{-5pt}
    }
    \subfloat[\small Overlap Ratio Trend\label{fig:overlap_trend}]{%
        \includegraphics[width=0.29\textwidth]{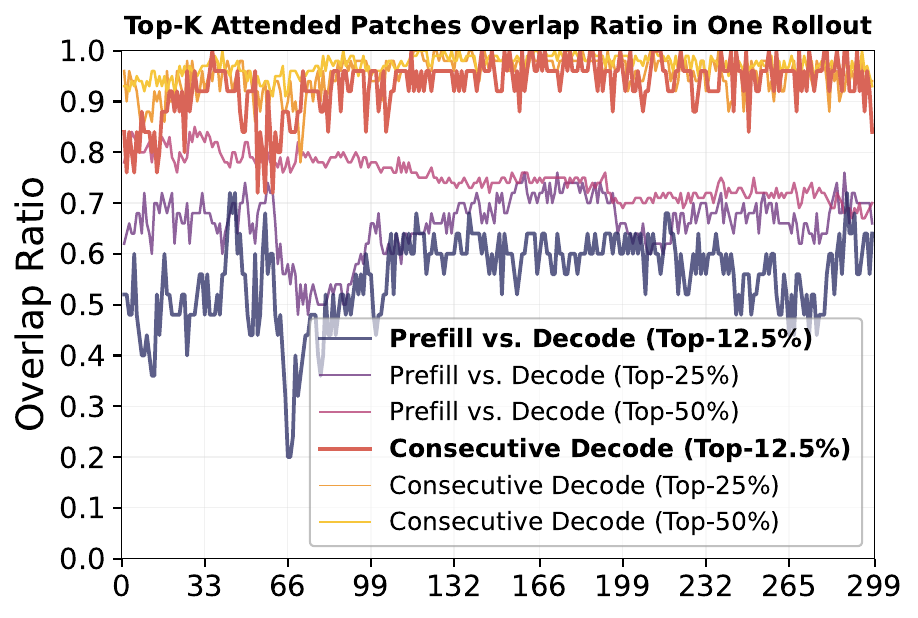}%
        \vspace{-3pt}
    }
    \hspace{0.01\textwidth}
    \subfloat[\small Prefill Attention\label{fig:prefill_attn}]{%
        \includegraphics[width=0.19\textwidth]{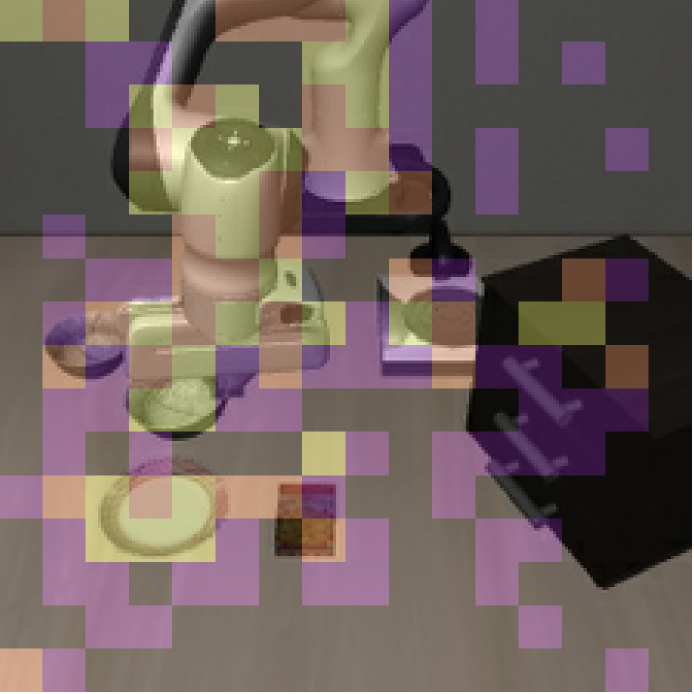}}
    \hspace{0.01\textwidth}
    \subfloat[\small Decode Attention\label{fig:decode_attn}]{%
        \includegraphics[width=0.19\textwidth]{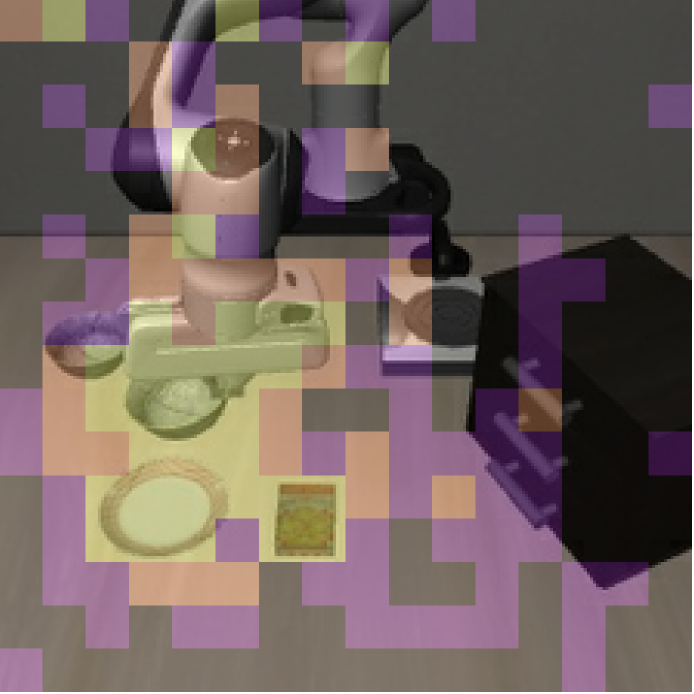}}
    \vspace{-0.2cm}
    \caption{\small
        \textbf{Distinct attention patterns across VLA inference.}
        (\textbf{a--b}) Overlap ratios of $\text{Top-}k$ attended patches among vision--language prefill, text-to-vision attention, action decode, and consecutive action-decode timesteps. We report average ratios (a) and one rollout trend (b).
        (\textbf{c--d}) Prefill (c) and action-decode (d) attention on the same frame, with top 12.5\% (yellow), 25\% (orange), and 50\% (purple) patches overlaid. Prefill attention shows broad semantic coverage, whereas action decoding is locally focused. These results reveal stage-wise visual-token importance mismatch in VLA inference.
        }
    \label{fig:dual_attention}
    \vspace{-0.5cm}
\end{figure*}

We systematically delve into the intrinsic characteristics of VLA inference (Figure~\ref{fig:dual_attention}). 
Our analysis proceeds in three stages. 
We first show that visual-token prefilling dominates the computational cost of VLA inference, making token reduction a direct route to acceleration (§\ref{sec:analysis_bottleneck}). 
We then reveal distinct attention patterns between semantic prefilling and action decoding, exposing the bias of semantic-only pruning criteria (§\ref{sec:analysis_attention}). 
Finally, we show that action-to-vision attention remains temporally consistent across consecutive control steps, enabling recent decode attention to approximate current action-level token importance (§\ref{sec:analysis_temporal}).
All analyses are conducted on OpenVLA~\citep{kim2024openvla} with the LIBERO benchmark~\citep{liu2023libero}. 
Together, these findings reveal a stage-wise mismatch in visual token importance: tokens that are salient for semantic prefilling are not necessarily sufficient for action decoding. 
Our findings identify the fundamental limitations of current VLM-specific token pruning methods for VLA models, which explains the performance degradation of semantic-only pruning methods and provides the empirical basis for VLA-Pruner.

\subsection{Visual-Token Prefilling as the Computational Bottleneck}
\label{sec:analysis_bottleneck}
We first characterize the computational bottleneck of VLA inference. 
As noted in prior VLM studies~\citep{zhang2024sparsevlm,chen2024fastv}, incorporating visual tokens inevitably introduces substantial memory and computational overhead. 
This overhead is further amplified in VLA models, which repeatedly process dense visual observations at every control step. 
In typical VLA settings, each frame typically contains $256 \times n$ visual tokens, where $n$ denotes the number of camera views, often an order of magnitude more than text tokens ($\sim$30--50) or action tokens ($\sim$7--56). 
Thus, processing the input visual stream becomes a primary contributor to inference cost.  
Since visual representations are highly redundant~\citep{alvar2025divprune,chen2024fastv,zhang2024sparsevlm}, visual token pruning shows great promise for VLA acceleration.

\subsection{Stage-wise Attention Divergence in VLA Inference}
\label{sec:analysis_attention}

\paragraph{Analysis Setup.}
VLA models integrate high-level semantic understanding and task planning with low-level action execution~\citep{firoozi2025foundationsmodel_survey,han2024dualvla,huang2025thinkact,sloman1996dual,cui2025openhelix}. We examine whether these two functional requirements are reflected in the attention behavior of VLA models. Specifically, we compare visual token attention in two stages of VLA inference: the vision-language prefill stage and the action decode stage. The prefill stage uses visual and language tokens to build the vision-language context, while the action-decode stage queries this context to generate executable actions. Attention scores are averaged across Transformer layers unless otherwise specified.
\vspace{-1em}
\paragraph{Results.}
Figure~\ref{fig:dual_attention} reveals a clear divergence between the two stages. 
Prefill attention exhibits a broad semantic distribution, while action-decode attention shows localized focus essential for precise motor control, as exemplified in Figure~\ref{fig:prefill_attn} and Figure~\ref{fig:decode_attn}.  Quantitatively, the overlap ratio of top-attended visual tokens between vision-language prefilling and action decoding averages $\sim$50\% and often falls below 30\% in individual rollouts (Figure~\ref{fig:dual_attention}).
This indicates that action decoding relies on a set of visual tokens that partially coincides with the tokens salient during semantic prefilling.
\vspace{-1em}
\paragraph{Key Finding.}
This stage-wise attention divergence exposes a fundamental limitation of VLM-oriented pruning methods. 
Existing methods such as FastV~\citep{chen2024fastv} and SparseVLM~\citep{zhang2024sparsevlm} typically rank visual tokens using semantic salience measured during context prefilling, such as prefill attention or text-to-vision attention. 
Such criteria can be effective for standard VLM tasks, but they are biased toward tokens useful for semantic prefilling. 
In VLA inference, this bias may prematurely remove tokens that are less salient during prefilling but crucial for action decoding. 
This failure mode becomes especially harmful at high pruning ratios, since removing even a small number of high-importance tokens can substantially degrade model performance~\citep{zhang2023h2o,ge2023model_tell}. 
A qualitative example is provided in Figure~\ref{fig:more_analysis}.

\subsection{Temporal Continuity of Action-to-Vision Attention}
\label{sec:analysis_temporal}

\paragraph{Practical Challenge.}
The above analysis suggests that robust VLA token pruning should preserve both semantic-level and action-level visual information. However, pruning must be performed during early-layer prefilling to reduce the dominant visual-token computation~\citep{chen2024fastv}. At this point, the current action-decode attention is not yet available. This creates a practical tension: action-level visual importance is necessary for reliable token selection, but it cannot be directly observed when pruning decisions are made.
\vspace{-1em}
\paragraph{Temporal Analysis.} VLA robot manipulation exhibits \textit{temporal continuity}~\citep{xu2025vlacache}: visual observations and action targets typically evolve smoothly over short horizons. We therefore analyze the overlap between top-attended visual tokens in consecutive action-decode steps. We observe that action-to-vision attention maps are highly consistent across adjacent timesteps. The top-attended visual tokens at timestep $t$ strongly overlap with those at timestep $t-1$, and this consecutive-decode overlap is substantially higher than the overlap between prefill attention and action-decode attention. Figure~\ref{fig:dual_attention} further shows averaged ratios and a representative rollout.

\paragraph{\textit{Principle for VLA Token Pruning.}} Our analysis shows that VLA token pruning faces a key challenge beyond standard VLM pruning: visual tokens important for semantic prefilling do not fully cover those needed for action decoding. 
As a result, pruning solely based on prefill-stage semantic salience can remove action-critical tokens, explaining why directly applying VLM-oriented pruning methods degrades VLA performance. 
Meanwhile, the temporal continuity of action-to-vision attention provides a practical way to estimate current action-level token importance from recent decode attentions. 
These findings motivate a VLA-specific pruning strategy, which accounts for both semantic relevance and temporally estimated action relevance when selecting visual tokens.

\section{Methodology}
\label{sec:methodology}
\subsection{Problem Formulation}
\vspace{-0.5em}
We formulate VLA token pruning as selecting a compact subset of visual tokens that preserves the model's action prediction. 
Given visual tokens $\mathbf{E}_v$ with $|\mathbf{E}_v|=M$ and a target budget $\tilde{M}<M$, the pruning function $f$ selects an index subset $\tilde{\mathcal{I}}\subseteq\{1,\ldots,M\}$ with $|\tilde{\mathcal{I}}|=\tilde{M}$, and returns
\[
\tilde{\mathbf{E}}_v
=
f(\mathbf{E}_v)
=
\{\mathbf{E}_v[i]: i\in\tilde{\mathcal{I}}\}.
\]
The goal is to minimize the discrepancy between the original and pruned VLA output distributions:
\begin{equation}
\label{eq:pruning_problem}
\abovedisplayskip=6pt minus 3pt
\min_{f} \, \mathcal{L}(\mathcal{P}, \tilde{\mathcal{P}}) \quad \text{s.t.} \quad |f(\mathbf{E}_v)| = \tilde{M}
\belowdisplayskip=6pt minus 3pt
\end{equation}
where $\mathcal{P}$$=$$P(A | \mathbf{E}_\tau, \mathbf{E}_v)$ and 
$\tilde{\mathcal{P}}$$=$$P(A | \mathbf{E}_\tau, f(\mathbf{E}_v))$. $P(|)$ denotes the conditional generation probability of VLA model. $\mathcal{L}$ measures the difference in the model’s output with and without pruning, and $\tilde{M}$ indicates the number of retained tokens. To reflect the stage-wise structure of VLA inference, we decompose $\mathcal{P}$ into vision-language prefill $P_{vl}$ and action decode $P_{act}$ based on hidden states $\mathbf{Z}_\tau,\mathbf{Z}_v$:
\[
\begin{aligned}
\mathcal{P} &= P_{\text{vl}}(\mathbf{Z}_\tau, \mathbf{Z}_v | \mathbf{E}_\tau, \mathbf{E}_v)\cdot P_{\text{act}}(A | \mathbf{Z}_\tau, \mathbf{Z}_v),\\
\tilde{\mathcal{P}} &= \tilde{P}_{\text{vl}}(\tilde{\mathbf{Z}}_\tau, f(\tilde{\mathbf{Z}}_v) | \mathbf{E}_\tau, f(\mathbf{E}_v))\cdot \tilde{P}_{\text{act}}(A | \tilde{\mathbf{Z}}_\tau, f(\tilde{\mathbf{Z}}_v)).
\end{aligned}
\]
Here, $f$ sparsifies both the input $\mathbf{E}_v$ and corresponding hidden states $\mathbf{Z}_v$. For VLA inference, relying on single-level objectives—either semantic-level $\mathcal{L}(P_{\text{vl}}, \tilde{P}_{\text{vl}})$ or action-level $\mathcal{L}(P_{\text{act}}, \tilde{P}_{\text{act}})$—fails to preserve the joint probability $\mathcal{P}$. To address this, we introduce VLA-Pruner.

\subsection{The Proposed VLA-Pruner}
\label{sec:vla-pruner}

\begin{wrapfigure}{r}{0.48\columnwidth}
\vspace{-0.75em}
    \centering
    \includegraphics[width=0.5\columnwidth]{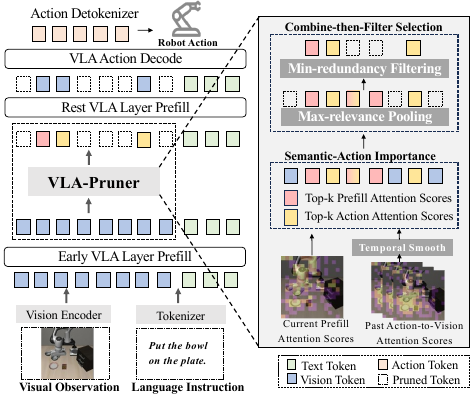}
    \vspace{-1.5em}
    \caption{\small \textbf{Overview of VLA-Pruner} (budget $k{=}3$). VLA-Pruner selects semantic- and action-salient tokens, then removes redundancy to preserve compact task-relevant information.}
    \label{fig:framework}
     \vspace{-1.75em}
\end{wrapfigure}

VLA-Pruner consists of \textit{semantic-action importance estimation} and a corresponding \textit{Combine-then-Filter token selection strategy} for VLA token pruning. The overall pipeline is illustrated in Figure~\ref{fig:framework} and detailed in Algorithm~\ref{alg:combine_filter}.

\subsubsection{Semantic-Action Importance Estimation}
\label{sec:dual importance}
For VLMs, previous works~\citep{chen2024fastv,zhang2024sparsevlm} have shown that prefill or text-to-vision attention scores can effectively measure visual-token importance. 
For VLA models, as discussed before (Sec.~\ref{sec:observation}), both semantic-level and action-level visual requirements should be satisfied. Therefore, VLA-Pruner estimates token importance from two complementary perspectives: prefill attention scores $\mathcal{S}^{t}_{\text{vl}}$ (Eq.~\ref{eq:prefill attention}) for semantic relevance, and action-to-vision attention scores $\mathcal{S}^{t}_{\text{act}}$ (Eq.~\ref{eq:action attention}) for action-level relevance.

\paragraph{Attention Estimation via Temporal Smoothing.}
As discussed in Sec.~\ref{sec:analysis_temporal}, action-to-vision attention exhibits short-term temporal continuity during VLA manipulation. 
However, the current action attention score $\mathcal{S}^{t}_{\text{act}}$ is unavailable when pruning decisions are made during prefilling. 
We therefore estimate current action-level importance from recent action-decode attentions $\{\mathcal{S}^{t-1}_{\text{act}}, \mathcal{S}^{t-2}_{\text{act}}, \dots\}$ using an Exponential Moving Average (EMA)-based temporal smoother~\citep{hyndman2018forecasting,gardner2006exponential,gardner1985exponentialsmmothing}. 
Specifically, since VLA manipulation depends more on short-term context~\citep{xu2025vlacache}, we use a finite-window EMA, implemented as a decaying window average over the previous $w$ steps:
\begin{equation}
\abovedisplayskip=6pt minus 2pt
\hat{\mathcal{S}}^{t}_{\text{act}} =
\frac{\sum_{i=1}^{w} \gamma^i \mathcal{S}^{t-i}_{\text{act}}}
{\sum_{i=1}^{w} \gamma^i},
\label{eq:our_temporal}
\belowdisplayskip=6pt minus 2pt
\end{equation}
where $w$ is the window size and $\gamma \in [0,1]$ controls the exponential decay rate. 
This estimator assigns larger weights to more recent action attentions while retaining short-term historical context, providing a simple proxy for current action-level visual importance. 
For brevity, we omit the time superscript $t$ in the following discussion.

\subsubsection{Semantic-Action Token Selection via Combine-then-Filter}
\label{sssec:dual selection}
Given the semantic importance score $\mathcal{S}_{\text{vl}}$ and temporally estimated action importance score $\hat{\mathcal{S}}_{\text{act}}$, a straightforward approach is to normalize them, take a weighted sum, and retain top-$\tilde{M}$ tokens. However, this score-fusion scheme has several limitations. First, it introduces a sensitive weighting hyperparameter, reducing robustness and generalizability. Second, it does not explicitly reduce redundancy, so similar or jointly moderate-salience tokens may dominate the retained set~\citep{alvar2025divprune}. Third, it relies on the estimated $\hat{\mathcal{S}}_{\text{act}}$, which can be unreliable under abrupt attention shifts.

To avoid these issues, VLA-Pruner follows a simple \textbf{Combine-then-Filter} strategy. This design follows the \textit{minimal-Redundancy-Maximal-Relevance} (mRMR) principle~\citep{peng2005featuremRMR}: we first preserve tokens relevant to either inference stage and then remove redundant candidates. We organize the procedure into three phases: \\
\noindent\textbf{(1) Semantic-Action Candidate Selection.}
We first select the top-$\tilde{M}$ visual tokens under the semantic and action importance scores, respectively:
\[
\mathcal{C}_{\text{vl}} =
\operatorname{Top-\tilde{M}}\left(\{\mathcal{S}_{\text{vl}}[i]\}_{i=1}^{M}\right),
\qquad
\mathcal{C}_{\text{act}} =
\operatorname{Top-\tilde{M}}\left(\{\hat{\mathcal{S}}_{\text{act}}[i]\}_{i=1}^{M}\right),
\]
where $\tilde{M}$ is the target token budget. 
Here, $\mathcal{C}_{\text{vl}}$ contains tokens salient for semantic prefilling, while $\mathcal{C}_{\text{act}}$ contains tokens estimated to be important for action decoding.\\
\noindent\textbf{(2) Relevance-Maximization Pooling.}
We then take the union of the two candidate sets:
\[
\mathcal{C}_{\text{dual}} = \mathcal{C}_{\text{vl}} \cup \mathcal{C}_{\text{act}}.
\]
This candidate pool preserves tokens relevant to either semantic prefilling or action decoding, maximizing two-stage relevance. 
Due to the attention divergence between these two stages, $|\mathcal{C}_{\text{dual}}|$ is generally larger than $\tilde{M}$.\\
\noindent\textbf{(3) Redundancy-Minimization Filtering.}
Finally, we filter redundant tokens from $\mathcal{C}_{\text{dual}}$ to meet the target budget $\tilde{M}$. Following diversity-based pruning~\citep{alvar2025divprune}, we formulate this step as a Max--Min Diversity Problem (MMDP)~\citep{porumbel2011mmdp}: selecting a subset $\tilde{\mathcal{C}} \subset \mathcal{C}_{\text{dual}}$ of size $\tilde{M}$ that maximizes the minimum pairwise distance among retained tokens:
\begin{equation}
\abovedisplayskip=6pt minus 3pt
\tilde{\mathcal{C}}
=
\argmax_{\mathcal{C} \subset \mathcal{C}_{\text{dual}},\, |\mathcal{C}|=\tilde{M}}
\;
\min_{\substack{i,j \in \mathcal{C}\\ i\neq j}}
d(v_i, v_j),
\label{eq:mmdp}
\belowdisplayskip=6pt minus 3pt
\end{equation}
where $v_i,v_j \in \mathbf{E}_v$ are visual patch embeddings indexed by $i,j$, and $d(\cdot,\cdot)$ is the cosine distance:
\begin{equation}
\abovedisplayskip=6pt minus 2pt
d(v_i, v_j)
=
1 -
\frac{v_i \cdot v_j}{\|v_i\|\,\|v_j\|}.
\belowdisplayskip=6pt minus 2pt
\label{eq:cosine}
\end{equation}
This objective selects a diverse subset from $\mathcal{C}_{\text{dual}}$, thereby reducing redundancy among retained tokens. MMDP can be solved by exact or heuristic methods~\citep{mmdpsolution2,marti2022mmdpsolution1}. To facilitate efficiency, we use a greedy algorithm that iteratively add the token whose minimum distance to the retained set is maximal, until reaching $\tilde{M}$ elements. To improve initialization, the first token is chosen by maximal second-nearest distance. The full procedure is detailed in Algorithm~\ref{alg:redundancy-minimization}.

\subsection{Implementation Details.}  
VLA-Pruner prunes visual tokens at Transformer layer K (K set to 3~\citep{chen2024fastv}). It selects tokens using last-layer prefill attention and temporal smoothing of action-to-vision attention averaged over the latter half layers. Pruned tokens are dropped from the rest prefill layers. 
For decaying window average, window size $w$ and decay rate $\gamma$ are empirically set as 3 and 0.8. We warm-start VLA-Pruner with $w$ steps to record sufficient attention history. VLA-Pruner is training-free, serving as a plug-and-play acceleration module for VLA architectures with action-to-vision cross-attention, a mechanism used by state-of-the-art VLA models~\citep{kim2024openvla,kim2025openvlaoft,black2024pi_0}, including (i) autoregressive policies~\citep{kim2024openvla} (e.g., OpenVLA~\citep{kim2024openvla}), (ii) action-chunk decoders (OpenVLA-OFT~\citep{kim2025openvlaoft}), and (iii) diffusion-head policies such as $\pi_0$~\citep{black2024pi_0}, for which we average action attention over flow-matching steps. We provide theoretical analysis of computational complexity in Appendix~\ref{app:flops}.

\section{Experiments}
\vspace{-1em}
\label{sec:experiments}
We evaluate VLA-Pruner in both simulation and real-world settings. We evaluate VLA-Pruner on open-source VLA models: OpenVLA~\citep{kim2024openvla}, OpenVLA-OFT~\citep{kim2025openvlaoft} and $\pi_0$~\citep{black2024pi_0}. All experiments are conducted on an NVIDIA RTX 4090 GPU.
\vspace{-0.75em}
\subsection{Experimental Setup}
\vspace{-0.5em}
\paragraph{Baselines and Evaluation Protocol.}
We compare VLA-Pruner with a comprehensive suite of training-free accelerators under given compute budgets. Baselines include state-of-the-art token pruning methods (FastV~\citep{chen2024fastv}, SparseVLM~\citep{zhang2024sparsevlm}, and DivPrune~\citep{alvar2025divprune}) and a VLA-specific method VLA-Cache~\citep{xu2025vlacache}. All methods are evaluated under identical budgets. To ensure a comprehensive comparison,  our experiments include diverse pruning ratios ({50\%, 75\%, 87.5\%}). For VLA-Cache, we set the same token reuse ratio. Our evaluation metrics mainly include: task success rate (\%), inference latency (ms) and FLOPs(T). We provide more details of baselines and VLA models in Appendix~\ref{app:setuo details}. EfficientVLA~\citep{yang2025efficientvla} uses token pruning strategies similar to prior work (using attention and diversity). We include results of it and HoloV~\citep{holov} in Appendix~\ref{app:more_baselines}.
\vspace{-1em}
\paragraph{Evaluation Benchmarks.}
We evaluate VLA-Pruner on simulation and real-robot benchmarks, with details in Appendix~\ref{app:benchmark details}.
\textbf{LIBERO:} LIBERO~\citep{liu2023libero} includes four suites—\textit{Spatial}, \textit{Object}, \textit{Goal}, and \textit{Long}—covering complementary manipulation generalization settings, with 10 tasks and 500 evaluation episodes per suite. Following OpenVLA~\citep{kim2024openvla} and OpenVLA-OFT~\citep{kim2025openvlaoft}, we use the standard evaluation protocol for direct comparison. We further evaluate the diffusion-head model $\pi_0$~\citep{black2024pi_0} to verify cross-architecture generality (Appendix~\ref{app:pi0_libero}).
\textbf{SIMPLER:} SIMPLER~\citep{li24simpler} provides two sim-to-real-oriented settings, visual matching (VM) and variant aggregation (VA). We report results on three tasks where OpenVLA achieves non-trivial success rates: \textit{Move Near}, \textit{Pick Coke Can} under VA, and \textit{Open/Close Drawer} under VM.
\textbf{Real Robot:} We evaluate VLA-Pruner on a 6-DoF xArm6 robot with a parallel gripper, using four real-world manipulation tasks (100 trials each): \textit{Can Stack}, \textit{Cup Pour}, \textit{Cube Place}, and \textit{Bottle Place}.

\label{sec:main results}
 \vspace{-1em}
\subsection{Main Results}
\begin{figure}[t]
\vspace{-1em}
    \centering
    \subfloat{%
        \includegraphics[width=0.245\textwidth]{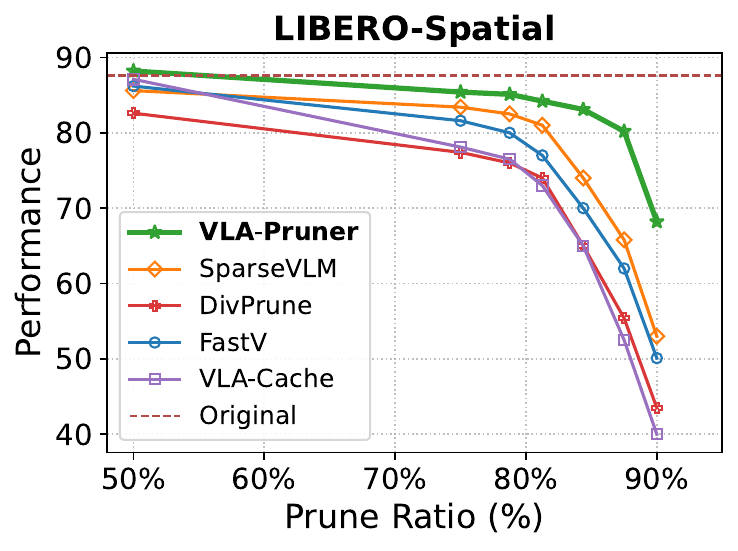}%
    }
    \subfloat{%
        \includegraphics[width=0.245\textwidth]{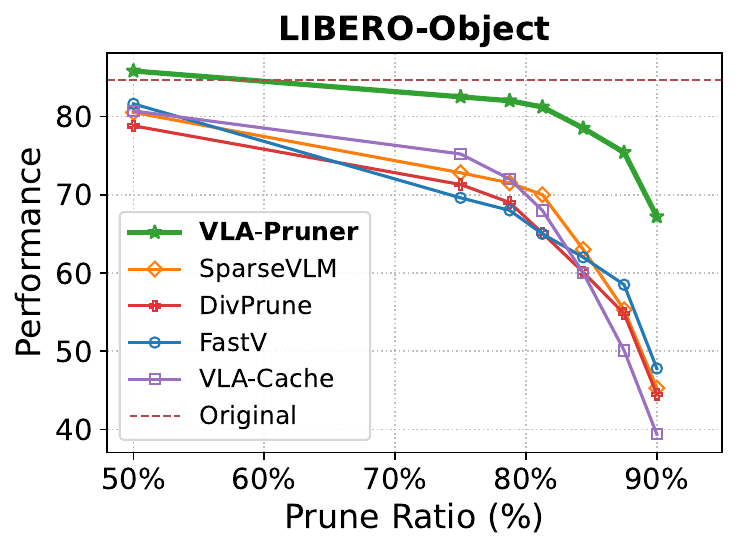}%
    }
    \subfloat{%
        \includegraphics[width=0.245\textwidth]{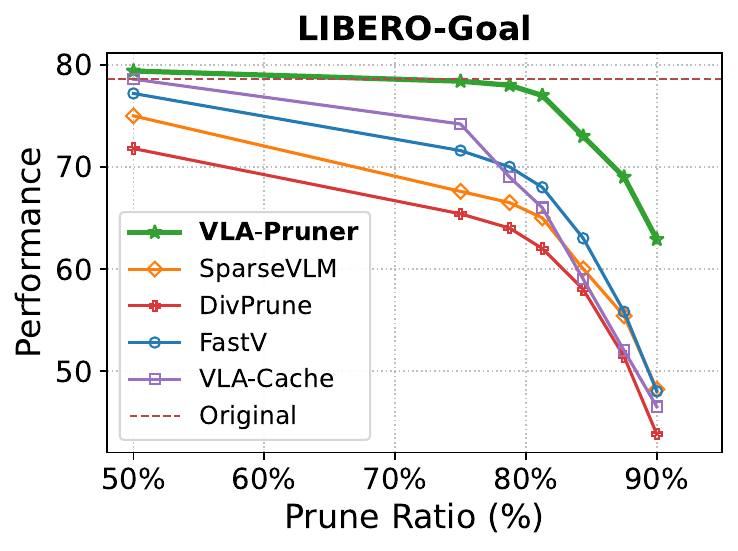}%
    }
    \subfloat{%
        \includegraphics[width=0.245\textwidth]{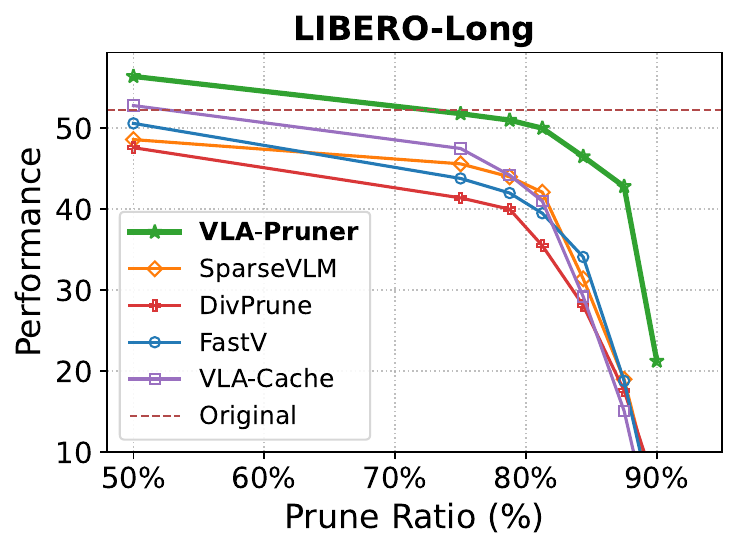}%
    }
    \vspace{-0.75em}
    \caption{\small \textbf{Performance of OpenVLA with baselines across LIBERO under varying pruning ratio.}}
    \label{fig:results}
    \vspace{-1.5em}
\end{figure}

\vspace{-0.75em}
\paragraph{Results on LIBERO.}
We evaluate VLA-Pruner on OpenVLA (fine-tuned) and OpenVLA-OFT on LIBERO benchmark. Table ~\ref{tab:main_results} reports \emph{success rate} on the four suites under token-retention ratios of 50\%, 25\%, and 12.5\%. We report a \emph{relative accuracy} metric, defined as the performance preservation ratio. VLA-Pruner achieves the best success rate and relative accuracy across all settings among training-free accelerators. Even at a 12.5\% retention ratio, it maintains 88.9\% and 88.27\% relative accuracy on OpenVLA and OpenVLA-OFT, respectively, outperforming baselines by up to 34.39\%. At a 50\% retention ratio, VLA-Pruner can improve success rate, thanks to precise noise filtering that stabilizes policy execution. This is evident on \textit{LIBERO-Long}, where it achieves 4.2\% improvement. DivPrune degrades VLA performance since manipulation needs localized visual details. VLA-Cache achieves competitive results due to temporal awareness but fails at high compression ratios. Figure~\ref{fig:results} visualizes more details of VLA-Pruner performance varying under different ratios.
\begin{table*}[t] \centering \tiny \caption{\footnotesize \textbf{Performance of VLA-Pruner on OpenVLA and OpenVLA-OFT across LIBERO} at 50\%, 25\%, and 12.5\% vision-token retention. For each method, the first line shows average success rates(\%); the second line shows relative accuracy vs.\ vanilla (100\%) or speedup ratio ($\times$). We also report FLOPs (T), inference latency (\(\mathrm{ms}/\mathrm{action}\); \(\mathrm{ms}/\mathrm{action\text{-}chunk}\)).} \vspace{-0.3cm} \makebox[\textwidth][c]{\resizebox{0.98\textwidth}{!}{ \centering \setlength{\tabcolsep}{3.0pt} \renewcommand{\arraystretch}{0.98} \begin{tabular}{l|cccc|ccc|cccc|ccc} \specialrule{0.2em}{0pt}{0pt} \multirow[c]{2}{*}{\textbf{Method}} & \multicolumn{7}{c|}{\textbf{OpenVLA}} & \multicolumn{7}{c}{\textbf{OpenVLA-OFT}} \\ \cline{2-8}\cline{9-15} & \textbf{Spatial} & \textbf{Object} & \textbf{Goal} & \textbf{Long} & \textbf{Acc.(\%)$\uparrow$} & \textbf{FLOPs(T)$\downarrow$} & \textbf{Latency(ms)$\downarrow$} & \textbf{Spatial} & \textbf{Object} & \textbf{Goal} & \textbf{Long} & \textbf{Acc.(\%)$\uparrow$} & \textbf{FLOPs(T)$\downarrow$} & \textbf{Latency(ms)$\downarrow$} \\ \hline \rowcolor{gray!12} \multicolumn{15}{c}{\textit{Upper Bound (100\%)}}\\ \hline \multirow[c]{2}{*}{Vanilla} & 87.6 & 84.6 & 78.6 & 52.2 & 100.0 & 1.906 & 236.41 & 95.8 & 98.7 & 96.3 & 90.7 & 100.0 & 3.903 & 135.78 \\ & 100\% & 100\% & 100\% & 100\% & (—) & 100.00\% & 1.00$\times$ & 100\% & 100\% & 100\% & 100\% & (—) & 100\% & 1.00$\times$ \\ \hline \rowcolor{gray!12} \multicolumn{15}{c}{\textit{Retain 50\% Tokens} \textcolor{MyGreen}{(\textbf{$\downarrow$ 50\%})}}\\ \hline \multirow[c]{2}{*}{FastV} & 86.2 & 81.6 & 77.2 & 50.6 & 97.43 & 1.136 & 172.32 & 94.6 & 96.8 & 92.7 & 87.0 & 97.26 & 2.219 & 88.23 \\ & 98.4\% & 96.5\% & 98.2\% & 96.9\% & ($\downarrow$ 2.57) & 59.60\% & 1.37$\times$ & 98.7\% & 98.1\% & 96.3\% & 95.9\% & ($\downarrow$ 2.74) & 56.85\% & 1.54$\times$ \\ \hline \multirow[c]{2}{*}{SparseVLM} & 85.6 & 80.5 & 75.0 & 48.6 & 95.44 & 1.155 & 175.77 & 94.1 & 93.7 & 91.2 & 85.1 & 95.37 & 2.289 & 90.22 \\ & 97.7\% & 95.2\% & 95.4\% & 93.1\% & ($\downarrow$ 4.56) & 60.60\% & 1.34$\times$ & 98.3\% & 94.9\% & 94.7\% & 93.8\% & ($\downarrow$ 4.63) & 58.65\% & 1.50$\times$ \\ \hline \multirow[c]{2}{*}{DivPrune} & 82.6 & 78.8 & 71.8 & 47.6 & 92.64 & 1.105 & 173.88 & 90.8 & 91.1 & 89.9 & 83.1 & 92.61 & 2.126 & 88.01 \\ & 94.3\% & 93.1\% & 91.3\% & 91.2\% & ($\downarrow$ 7.36) & 57.97\% & 1.36$\times$ & 94.8\% & 92.3\% & 93.3\% & 91.7\% & ($\downarrow$ 7.39) & 54.47\% & 1.54$\times$ \\ \hline \multirow[c]{2}{*}{VLA-Cache} & 87.1 & 80.7 & 78.6 & 51.8 & 98.48 & 1.384 & 192.20 & 95.4 & 96.0 & 96.7 & 90.2 & 99.09 & 2.730 & 101.01 \\ & 99.4\% & 95.4\% & 100.0\% & 99.2\% & ($\downarrow$ 1.52) & 72.61\% & 1.23$\times$ & 99.6\% & 97.3\% & 100\% & 99.4\% & ($\downarrow$ 0.91) & 69.95\% & 1.34$\times$ \\ \hline \multirow[c]{2}{*}{\textbf{VLA-Pruner}} & 88.2 & 85.8 & 79.4 & 56.4 & \textbf{102.45} & 1.139 & 178.38 & 97.3 & 98.6 & 96.8 & 92.6 & \textbf{101.05} & 2.234 & 92.86 \\ & 100.7\% & 101.4\% & 101.0\% & 108.0\% & \textcolor{MyGreen}{($\uparrow$ \textbf{2.45})} & 59.76\% & 1.33$\times$ & 101.6\% & 99.9\% & 100.5\% & 102.1\% & \textcolor{MyGreen}{($\uparrow$ \textbf{1.05})} & 57.24\% & 1.46$\times$ \\ \hline \rowcolor{gray!12} \multicolumn{15}{c}{\textit{Retain 25\% Tokens} \textcolor{MyGreen}{(\textbf{$\downarrow$ 75\%})}}\\ \hline \multirow[c]{2}{*}{FastV} & 81.6 & 69.6 & 71.6 & 43.8 & 87.62 & 0.757 & 141.25 & 87.8 & 81.8 & 87.6 & 74.6 & 87.48 & 1.401 & 72.73 \\ & 93.2\% & 82.3\% & 91.1\% & 83.9\% & ($\downarrow$ 12.38) & 39.72\% & 1.67$\times$ & 91.7\% & 82.9\% & 91.0\% & 82.3\% & ($\downarrow$ 12.52) & 35.89\% & 1.87$\times$ \\ \hline \multirow[c]{2}{*}{SparseVLM} & 83.4 & 72.8 & 67.6 & 45.6 & 88.67 & 0.772 & 144.12 & 89.8 & 84.8 & 82.7 & 79.1 & 88.63 & 1.459 & 74.71 \\ & 95.2\% & 86.1\% & 86.0\% & 87.4\% & ($\downarrow$ 11.33) & 40.50\% & 1.64$\times$ & 93.7\% & 85.9\% & 85.9\% & 87.2\% & ($\downarrow$ 11.37) & 37.38\% & 1.82$\times$ \\ \hline \multirow[c]{2}{*}{DivPrune} & 77.4 & 61.3 & 65.4 & 41.4 & 80.96 & 0.743 & 142.12 & 83.9 & 72.6 & 79.3 & 72.5 & 80.76 & 1.389 & 72.30 \\ & 88.4\% & 72.5\% & 83.2\% & 79.3\% & ($\downarrow$ 19.04) & 38.98\% & 1.66$\times$ & 87.6\% & 73.6\% & 82.3\% & 79.9\% & ($\downarrow$ 19.24) & 35.58\% & 1.88$\times$ \\ \hline \multirow[c]{2}{*}{VLA-Cache} & 78.1 & 73.2 & 70.2 & 45.5 & 88.08 & 0.961 & 164.17 & 85.6 & 85.3 & 84.1 & 80.3 & 88.11 & 1.938 & 85.02 \\ & 89.2\% & 86.5\% & 89.3\% & 87.2\% & ($\downarrow$ 11.92) & 50.42\% & 1.44$\times$ & 89.4\% & 86.4\% & 87.3\% & 88.5\% & ($\downarrow$ 11.89) & 49.65\% & 1.60$\times$ \\ \hline \multirow[c]{2}{*}{\textbf{VLA-Pruner}} & 85.4 & 82.5 & 78.4 & 51.8 & \textbf{98.48} & 0.758 & 144.81 & 93.5 & 96.2 & 95.2 & 90.2 & \textbf{98.37} & 1.420 & 75.64 \\ & 97.5\% & 97.5\% & 99.7\% & 99.2\% & \textcolor{MyGreen}{($\downarrow$ \textbf{1.52})} & 39.77\% & 1.63$\times$ & 97.6\% & 97.5\% & 98.9\% & 99.5\% & \textcolor{MyGreen}{($\downarrow$ \textbf{1.63})} & 36.38\% & 1.80$\times$ \\ \hline \rowcolor{gray!12} \multicolumn{15}{c}{\textit{Retain 12.5\% Tokens} \textcolor{MyGreen}{(\textbf{$\downarrow$ 87.5\%})}}\\ \hline \multirow[c]{2}{*}{FastV} & 62.0 & 58.5 & 55.8 & 18.8 & 63.08 & 0.568 & 125.76 & 60.6 & 59.2 & 60.6 & 28.6 & 61.64 & 1.082 & 66.11 \\ & 70.8\% & 69.1\% & 71.0\% & 36.0\% & ($\downarrow$ 36.92) & 29.80\% & 1.88$\times$ & 63.3\% & 60.0\% & 62.9\% & 31.5\% & ($\downarrow$ 38.36) & 27.72\% & 2.05$\times$ \\ \hline \multirow[c]{2}{*}{SparseVLM} & 65.8 & 55.3 & 55.4 & 19.0 & 63.20 & 0.581 & 128.77 & 73.1 & 65.0 & 67.5 & 32.8 & 62.11 & 1.136 & 67.65 \\ & 75.1\% & 65.4\% & 70.5\% & 36.4\% & ($\downarrow$ 36.80) & 30.48\% & 1.84$\times$ & 76.3\% & 65.8\% & 70.1\% & 36.1\% & ($\downarrow$ 37.89) & 29.11\% & 2.01$\times$ \\ \hline \multirow[c]{2}{*}{DivPrune} & 55.4 & 54.8 & 51.4 & 17.6 & 58.06 & 0.569 & 125.36 & 61.4 & 61.8 & 64.9 & 30.8 & 56.83 & 1.059 & 66.46 \\ & 63.2\% & 64.8\% & 65.4\% & 33.7\% & ($\downarrow$ 41.94) & 29.28\% & 1.89$\times$ & 64.1\% & 62.6\% & 67.4\% & 34.0\% & ($\downarrow$ 43.17) & 27.13\% & 2.04$\times$ \\ \hline \multirow[c]{2}{*}{VLA-Cache} & 52.5 & 50.1 & 52.0 & 15.1 & 54.79 & 0.710 & 145.93 & 57.3 & 58.7 & 64.8 & 26.2 & 53.88 & 1.373 & 76.15 \\ & 59.9\% & 59.2\% & 66.2\% & 28.9\% & ($\downarrow$ 45.21) & 37.25\% & 1.62$\times$ & 59.8\% & 59.5\% & 67.3\% & 28.9\% & ($\downarrow$ 46.12) & 35.18\% & 1.78$\times$ \\ \hline \multirow[c]{2}{*}{\textbf{VLA-Pruner}} & 80.2 & 78.4 & 69.0 & 42.8 & \textbf{88.90} & 0.571 & 129.01 & 88.1 & 87.6 & 84.9 & 68.8 & \textbf{88.27} & 1.096 & 68.95 \\ & 91.6\% & 92.7\% & 87.8\% & 82.0\% & \textcolor{MyGreen}{($\downarrow$ \textbf{11.10})} & 29.96\% & 1.83$\times$ & 91.9\% & 91.7\% & 88.1\% & 81.3\% & \textcolor{MyGreen}{($\downarrow$ \textbf{11.73})} & 28.08\% & 1.99$\times$ \\ \specialrule{0.2em}{0pt}{0pt} \end{tabular} } } \label{tab:main_results} \vspace{-2em} \end{table*}

\begin{wraptable}{r}{0.48\textwidth}
\centering
\vspace{-0.75em}
\setlength{\tabcolsep}{3pt}
\renewcommand{\arraystretch}{0.88}
\footnotesize
\caption{\small \textbf{SIMPLER results at 75\% pruning ratio.}}
\vspace{-0.75em}
\begin{adjustbox}{width=\linewidth,keepaspectratio=false}
\begin{tabular}{l|c|c|c|c} 
\specialrule{0.18em}{0pt}{0pt}
\textbf{Method} 
& \makecell{\textbf{Move}\\\textbf{Near}} 
& \makecell{\textbf{Pick}\\\textbf{Coke Can}} 
& \makecell{\textbf{Open/Close}\\\textbf{Drawer}} 
& \textbf{Overall} \\
\hline
\multirow{2}{*}{OpenVLA}
  & 54.0\% & 52.8\% & 49.5\% & 52.1\% \\
  & (100\%) & (100\%) & (100\%) & (100\%) \\
\hline
\multirow{2}{*}{FastV}
  & 38.7\% & 41.9\% & 33.7\% & 38.1\% \\
  & (71.7\%) & (79.4\%) & (68.1\%) & (73.1\%) \\
\hline
\multirow{2}{*}{VLA-Cache}
  & 41.2\% & 40.6\% & 38.8\% & 40.2\% \\
  & (76.3\%) & (76.9\%) & (78.4\%) & (77.2\%) \\
\hline
\multirow{2}{*}{\bf VLA-Pruner}
  & 52.4\% & 50.1\% & 48.8\% & 50.4\% \\
  & (\textbf{97.0\%}) & (\textbf{94.9\%}) & (\textbf{98.6\%}) & (\textbf{96.8\%}) \\
\specialrule{0.18em}{0pt}{0pt}
\end{tabular}
\end{adjustbox}
\vspace{-1.25em}
\label{tab:SIMPLER}
\end{wraptable}
\vspace{-1em}
\paragraph{Results on SIMPLER.}
We evaluate cross-environment generalization on the SIMPLER environment. Table~\ref{tab:SIMPLER} compares VLA-Pruner with acceleration baselines on OpenVLA at 75\% compression ratio. VLA-Pruner best preserves accuracy, showing robust transfer across environments.

\begin{figure}[t]
  \centering
  \setlength{\abovecaptionskip}{2pt}
  \setlength{\belowcaptionskip}{-6pt}

  \begin{minipage}[t]{0.50\linewidth}
    \vspace{0pt}
    \centering
    \includegraphics[width=\linewidth]{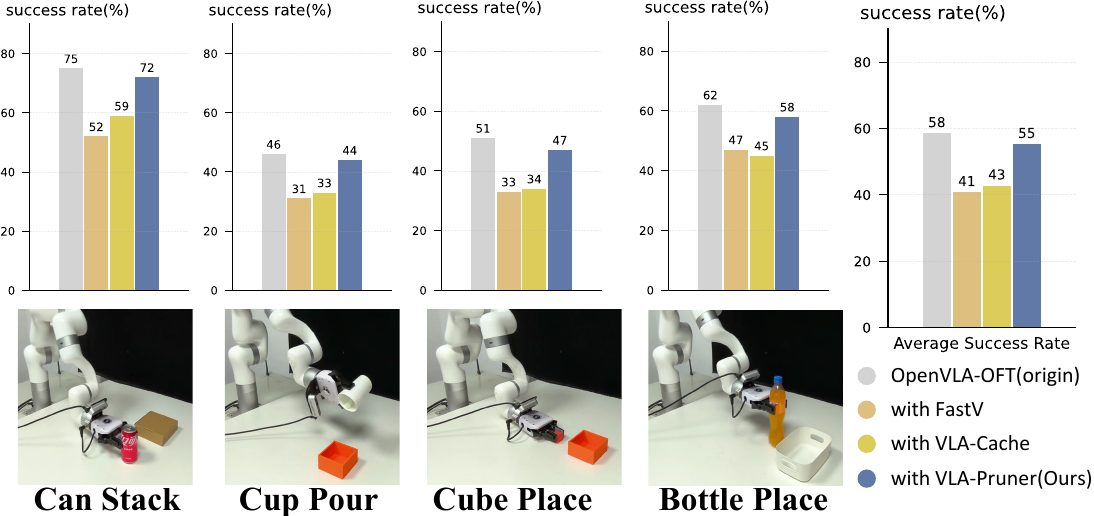}
    \caption{\small \textbf{Performance of VLA-Pruner on OpenVLA-OFT for real-robot tasks under a 75\% pruning ratio.}
    We use a 6-DoF xArm6 robotic arm. VLA-Pruner better preserves model performance, demonstrating its practical advantage.}
    \label{fig:realworld}
    \vspace{-1em}
  \end{minipage}
  \hfill
  \begin{minipage}[t]{0.48\linewidth}
    \vspace{0pt}
    \centering

    \makebox[\linewidth][c]{%
      \includegraphics[width=0.49\linewidth]{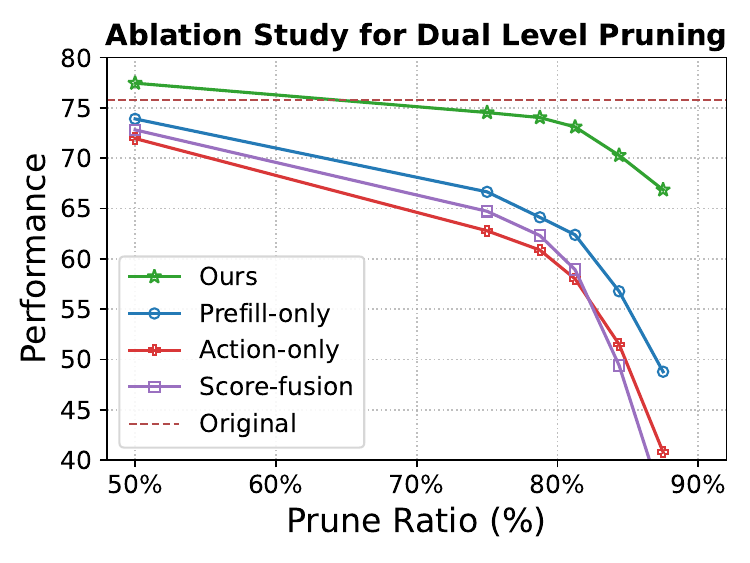}%
      \hspace{0.01\linewidth}%
      \includegraphics[width=0.49\linewidth]{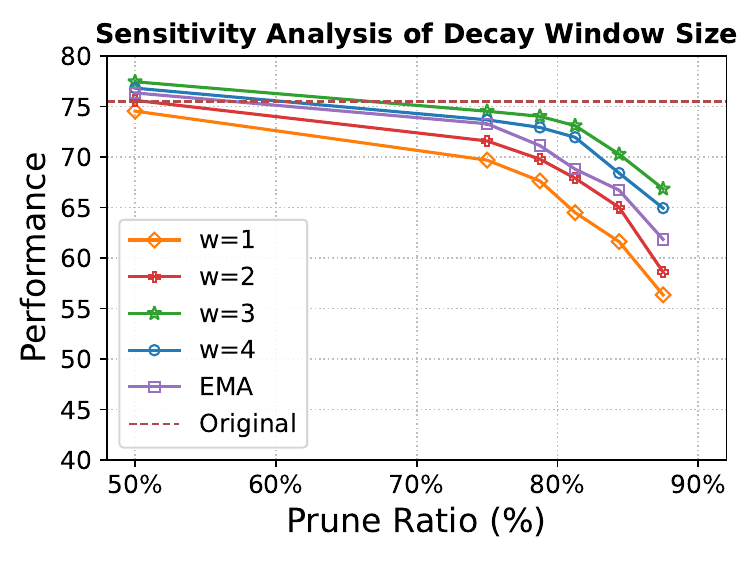}%
    }
    \vspace{0.25em}
    \includegraphics[width=\linewidth]{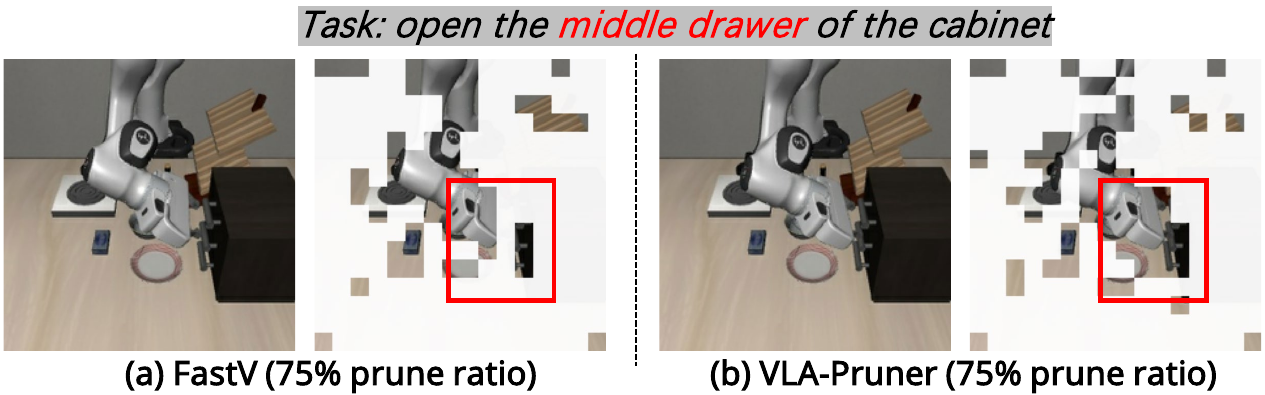}
    \caption{\small \textbf{Analysis of VLA-Pruner.}}
    \label{fig:more_analysis}
    \vspace{-1em}
  \end{minipage}
\end{figure}

\vspace{-1em}
\paragraph{Results on Real Robot.}
We evaluate real-world performance on a physical 6-DoF xArm6 arm controlled by OpenVLA-OFT. Figure~\ref{fig:realworld} reports results under 75\% token-pruning ratio. VLA-Pruner attains the highest relative accuracy, underscoring its practicality for on-robot deployment.
\vspace{-1em}
\paragraph{Efficiency Analysis.}
We report inference efficiency metrics in Table~\ref{tab:main_results}, including inference latency and FLOPs(T). At same pruning ratios, VLA-Pruner achieves comparable speedups to visual token–pruning methods. Its latency is $\sim$5 ms higher than FastV, yet it achieves significantly better performance. At the 25\% retention ratio, VLA-Pruner outperforms token pruning baselines at 50\% retention, while being $\sim$16\% faster and using $\sim$20\% fewer FLOPs. VLA-Pruner is more compute-efficient than VLA-Cache. VLA-Pruner requires storing historical attention, but it introduces a negligible increase in Max GPU memory (see Appendix~\ref{app:more_results:memory}).

\subsection{Analysis of VLA-Pruner}
\vspace{-0.75em}
\paragraph{Ablation Studies.}
We evaluate three alternative token selection criteria: (1) \textbf{\textit{prefill-only}}, relying solely on prefill attention; (2) \textbf{\textit{action-only}}, utilizing only action decode attention; and (3) \textbf{\textit{score-fusion}}, a naive weighted linear combination of both. As shown in Figure~\ref{fig:more_analysis}, VLA-Pruner performs best, validating the effectiveness of our token selection strategy.
\vspace{-1.5em}
\paragraph{Sensitivity Analysis.}
We analyze the sensitivity of VLA-Pruner to the decay window size $w$ in Figure~\ref{fig:more_analysis}, which demonstrates the robustness of our method. We further study varying decay rate $\gamma$ and pruning layer $K$ in Appendix~\ref{app:more_results:ablations}.
\vspace{-1em}
\paragraph{Qualitative Visualization.}
We provide a qualitative case study in Figure~\ref{fig:more_analysis}. VLA-Pruner maintains broad semantic understanding and preserves essential localized visual information for action. 
\vspace{-0.25em}
\section{Related Works}
\vspace{-0.75em}
\paragraph{Vision-Language-Action Models (VLA)}
Large-scale VLMs have advanced multimodal learning by integrating visual perception with language reasoning~\citep{li2023vision}. Building on this progress, VLA models~\citep{kim2024openvla,black2024pi_0,kim2025openvlaoft,brohan2022rt,zitkovich2023rt} further introduce an action modality for end-to-end motor control. They typically adopt large VLM backbones~\citep{touvron2023llama2}, fine-tune them on robotic data~\citep{o2024openx-embodiment}, and generate actions either by discretizing them into language-like tokens, as in OpenVLA~\citep{kim2024openvla}, or by attaching diffusion-policy heads, as in $\pi_0$~\citep{black2024pi_0}. In both cases, fine-grained action generation commonly relies on \textit{action-to-vision cross-attention}. Despite strong performance on robot manipulation benchmarks~\citep{liu2023libero,li24simpler}, VLA models remain computationally intensive, hindering real-time deployment on resource-constrained platforms.
\vspace{-2em}
\paragraph{Visual Token Pruning for VLMs}
Visual token pruning~\citep{chen2024fastv,zhang2024sparsevlm,alvar2025divprune,ye2025fitprune} is a widely adopted strategy to reduce visual redundancy in VLMs. One line of these methods defines \textit{semantic importance criteria} to rank and prune visual tokens, e.g., FastV~\citep{chen2024fastv} (using prefill attention), SparseVLM~\citep{zhang2024sparsevlm} (using text–to-vision attention), DivPrune~\citep{alvar2025divprune} (using diversity-driven selection), and HoloV~\citep{holov} (using crop-wise attention and diversity). Another line of \emph{calibration-based} approaches (e.g., FitPrune~\citep{ye2025fitprune}, VTW~\citep{lin2025VTW}) select pruning layers/ratios by analyzing model outputs on a calibration set. However, existing methods either remain driven by semantic salience or require costly calibration, failing to explicitly account for the action-dependent visual requirements of VLA inference. This motivates a VLA-specific pruning strategy.
\vspace{-1em}
\paragraph{Training-Free Acceleration for VLA Models}
Visual token pruning has recently been integrated into training-free VLA acceleration frameworks~\citep{surveyefficientvla}. To name a few, EfficientVLA~\citep{surveyefficientvla}, SP-VLA~\citep{li2025sp}, and SpecPrune-VLA~\citep{wang2025specprune} perform structured acceleration that incorporates token pruning. However, these methods still use semantic salience to retain tokens. Thus, they rely on additional modules (e.g., diffusion-feature cache~\citep{yang2025efficientvla}, lightweight generation module~\citep{li2025sp}, or layer-reduction~\citep{yang2025efficientvla,wang2025specprune}) to balance performance and speed, limiting generalizability and effectiveness. VLA-Cache~\citep{xu2025vlacache} exploits temporal continuity by caching static visual token features, opening a promising direction. However, its cache-based mechanism is less efficient.  

\section{Conclusion}
\vspace{-1em}
We propose VLA-Pruner, a training-free token pruning method for efficient VLA inference. We reveal a stage-wise mismatch in VLA visual-token importance and addresses this by combining semantic relevance with temporally estimated action relevance. VLA-Pruner selects compact, non-redundant visual tokens through a Combine-then-Filter strategy. Across diverse manipulation tasks and multiple VLA architectures, VLA-Pruner reduces computation while preserving performance.

\bibliographystyle{plainnat}
\bibliography{ref}

\clearpage
\appendix

\section{Method Details}
\label{app:method details}
\subsection{Algorithm Outlines}
\label{app:algorithms}
We detail VLA-Pruner in Algorithm~\ref{alg:combine_filter} (overall pipeline) and Algorithm~\ref{alg:redundancy-minimization} (min-redundancy filtering).

\begin{algorithm}[tb]
\caption{Overall Pipeline of VLA-Pruner}
\label{alg:combine_filter}
\small
\begin{algorithmic}[1]
\STATE {\bfseries Input:} Visual embeddings $\mathbf{E}_v$, Semantic attention scores $\mathcal{S}_{\mathrm{vl}}$, Action attention history $\{\mathcal{S}_{\text{act}}^{t-i}\}_{i=1}^w$, Token budget $\tilde{M}$, Decay params $\gamma$
\STATE {\bfseries Output:} Selected index set $\tilde{\mathcal{C}} \subseteq \{1,\dots,M\}$, with $|\tilde{\mathcal{C}}| = \tilde{M}$

\IF{$\tilde{M} \ge M$}
    \STATE {\bfseries return} $\{1,\dots,M\}$ \BlueComment{Skipping pruning}
\ENDIF
\STATE $\hat{\mathcal{S}}_{\text{act}} \leftarrow \sum_{i=1}^{w} \gamma^i \mathcal{S}^{t-i}_{\text{act}} / \sum_{i=1}^{w} \gamma^i$ \BlueComment{Action Attention Estimation via Temporal Smoothing (Eq.~\ref{eq:our_temporal})}
\STATE $\mathcal{C}_{\mathrm{vl}} \gets \text{indices of Top-}\tilde{M}\text{ elements of } \mathcal{S}_{\mathrm{vl}}$
\STATE $\mathcal{C}_{\mathrm{act}} \gets \text{indices of Top-}\tilde{M}\text{ elements of } \hat{\mathcal{S}}_{\text{act}}$
\STATE \textbf{Max-Relevance pooling:} $\mathcal{C}_{\mathrm{dual}} \gets \mathcal{C}_{\mathrm{vl}} \cup \mathcal{C}_{\mathrm{act}}$
\STATE \textbf{Min-Redundancy filtering:} $\tilde{\mathcal{C}} \gets \textsc{Redundancy-Minimization Filtering}(\mathbf{E}_v, \mathcal{C}_{\mathrm{dual}}, \tilde{M})$ \BlueComment{Alg.~\ref{alg:redundancy-minimization}}

\STATE {\bfseries return} $\tilde{\mathcal{C}}$
\end{algorithmic}
\end{algorithm}
\vspace{-0.4cm}

\begin{algorithm}[tb]
\caption{Redundancy-Minimization Filtering}
\label{alg:redundancy-minimization}
\small
\setlength{\lineskip}{0pt}
\setlength{\lineskiplimit}{0pt}
\begin{algorithmic}[1]
\STATE {\bfseries Input:} Visual embeddings $\mathbf{E}_v = \{v_1,\dots,v_M\}$, candidate index set $\mathcal{C}_{\mathrm{dual}} \subseteq \{1,\dots,M\}$, target size $\tilde{M}$
\STATE {\bfseries Output:} Selected index set $\tilde{\mathcal{C}}$, with $|\tilde{\mathcal{C}}| = \tilde{M}$

\IF{$|\mathcal{C}_{\mathrm{dual}}| \le \tilde{M}$}
    \STATE {\bfseries return} $\mathcal{C}_{\mathrm{dual}}$
\ENDIF

\STATE Extract candidate features: $\{v_i\}_{i \in \mathcal{C}_{\mathrm{dual}}}$
\STATE Normalize: $u_i \gets v_i / \|v_i\|_2$ for all $i \in \mathcal{C}_{\mathrm{dual}}$
\STATE Compute cosine distance matrix $D$ over $\mathcal{C}_{\mathrm{dual}}$: $D_{ij} \gets 1 - u_i^\top u_j$

\STATE $\tilde{\mathcal{C}} \gets \emptyset$ \BlueComment{Initialize by maximal second-nearest distance}

\FOR{each $i \in \mathcal{C}_{\mathrm{dual}}$}
    \STATE Let $\{d_{i1}, d_{i2}, \dots\}$ be $D_{ij}$ for $j \in \mathcal{C}_{\mathrm{dual}}, j \neq i$ sorted increasingly
    \STATE $s_i \gets d_{i2}$ \BlueComment{Second smallest distance}
\ENDFOR

\STATE $i^\star \gets \arg\max_{i \in \mathcal{C}_{\mathrm{dual}}} s_i$
\STATE $\tilde{\mathcal{C}} \gets \tilde{\mathcal{C}} \cup \{i^\star\}$

\WHILE{$|\tilde{\mathcal{C}}| < \tilde{M}$}
    \FOR{each $j \in \mathcal{C}_{\mathrm{dual}} \setminus \tilde{\mathcal{C}}$}
        \STATE $m_j \gets \min_{i \in \tilde{\mathcal{C}}} D_{ij}$ \BlueComment{Min distance to selected set}
    \ENDFOR
    \STATE $j^\star \gets \arg\max_{j \in \mathcal{C}_{\mathrm{dual}} \setminus \tilde{\mathcal{C}}} m_j$
    \STATE $\tilde{\mathcal{C}} \gets \tilde{\mathcal{C}} \cup \{j^\star\}$
\ENDWHILE

\STATE {\bfseries return} $\tilde{\mathcal{C}}$
\end{algorithmic}
\end{algorithm}

\subsection{Complexity Analysis}
\label{app:flops}

Following FastV~\citep{chen2024fastv}, SparseVLM~\citep{zhang2024sparsevlm} and  DivPrune~\citep{alvar2025divprune}, we estimate the computational requirement of VLA-Pruner in terms of Transformer FLOPs, and focus on how pruning visual tokens changes the effective sequence length seen by each layer. As in prior work~\citep{alvar2025divprune}, we count the FLOPs of multi-head attention and feed-forward networks (FFNs) in the backbone (the overhead of token selection as negligible compared to the main model FLOPs.)
Let $T$ be the total number of Transformer layers in the VLA backbone and $K$ the index of the pruning layer (we prune tokens before the $K$ layer). Denote the numbers of text and visual tokens before pruning by $N$ and $M$ ($M \gg N$), and let $\mu = N + M$ be the full sequence length. After VLA-Pruner keeps a fraction $\rho$ of the visual tokens, the sequence length becomes $\tilde{\mu} = N + \rho M$. Following~\citep{chen2024fastv}, the FLOPs of attention and FFN for one layer with sequence length $n$, hidden size $d$, and FFN intermediate size $m$ can be approximated as
\begin{equation}
C(n) = 4 n d^{2} + 2 n^{2} d + 2 n d m.
\end{equation}
Without pruning, the total FLOPs are
\begin{equation}
\text{FLOPs}_{\text{full}} = T \, C(\mu).
\end{equation}
With VLA-Pruner, the first $K-1$ layers still use the full sequence $\mu$, while the remaining $T-K+1$ layers use the shortened sequence $\tilde{\mu}$, giving
\begin{equation}
\text{FLOPs}_{\text{prune}} = (K-1) \, C(\mu) + (T-K+1) \, C(\tilde{\mu}).
\end{equation}
Thus, the theoretical FLOPs ratio is
\begin{equation}
\small
r_{\text{FLOPs}}
= \frac{\text{FLOPs}_{\text{prune}}}{\text{FLOPs}_{\text{full}}}
= \frac{(K-1) C(\mu) + (T-K+1) C(\tilde{\mu})}{T C(\mu)}.
\label{eq:our_flops_ratio}
\end{equation}
Since the attention term in $C(\cdot)$ is quadratic in sequence length, the reduction scales roughly with $\rho^2$ in layers after $K$. This matches the empirical FLOPs ratios in Tab.~\ref{tab:main_results}: VLA-Pruner reduces FLOPs to about $60\%$, $40\%$, and $30\%$ of the original model at $50\%$, $25\%$, and $12.5\%$ token retention, respectively.
Temporal smoothing over a short window of action-to-vision attention has complexity $O(wM)$ per timestep, and the greedy Max--Min filtering over the dual-level candidate pool of size $O(\rho M)$ is $O((\rho M)^2)$ per prefill. In practice, these components contribute little to the total wall-clock time, as these operations are invoked only once during the prefill stage.
Finally, Eq.~\eqref{eq:our_flops_ratio} shows that VLA-Pruner offers a flexible way to trade accuracy for efficiency: the retained-token ratio $\rho \in (0,1]$. Smaller $\rho$ reduces FLOPs quadratically in layers after $K$, yielding higher speedup, while larger $\rho$ recovers the original model's performance. In our experiments, we report results at $\rho \in {0.5, 0.25, 0.125}$, but VLA-Pruner can be used at any intermediate ratio to meet a given compute or latency target.

\section{Experiments Details}
\label{app:experiments details}

\subsection{Setting up Details}
\label{app:setuo details}
\subsubsection{VLA Model Details}
\label{app:vla_models}
We consider three representative VLA architectures in our experiments: OpenVLA, OpenVLA-OFT, and $\pi_0$. All models take RGB observations and a natural language instruction as input, and produce robot actions end-to-end.\\
\vspace{-0.05cm}
\noindent\textbf{OpenVLA.}
OpenVLA~\citep{kim2024openvla} is a 7B-parameter open-source vision–language–action model built by fine-tuning a Prismatic VLM (Llama~2 language backbone with DINOv2 and SigLIP visual encoders) on robot manipulation episodes from the Open X-Embodiment dataset. The model represents actions as discrete tokens and decodes them autoregressively, enabling a single policy to control multiple embodiments out-of-the-box. OpenVLA serves as our base autoregressive VLA, and we follow its official checkpoints and evaluation settings on LIBERO.\\
\noindent\textbf{OpenVLA-OFT.}
OpenVLA-OFT~\citep{kim2025openvlaoft} is an “Optimized Fine-Tuning” variant of OpenVLA that improves both control performance and inference efficiency. It augments the original decoder with parallel decoding and action chunking, and switches to a continuous action representation trained with a simple L1 regression objective. This substantially boosts success rates on LIBERO while increasing action-generation throughput. To be noted, OpenVLA-OFT utilizes bidirectional attention, meaning that during inference, there is already an action-to-vision cross-attention. However, we observe that during the early-layer prefill phase, the attention still focuses on semantic understanding, making action-to-vision less significant. In deeper layers, the action-to-vision attention starts to exhibit more distinct properties. This allows our method to effectively enhance token pruning performance by utilizing past timestep action-to-vision cross-attention at the deeper layers to guide current token pruning.\\
\noindent\textbf{$\pi_0$ Model.}
$\pi_0$~\citep{black2024pi_0} is a prototype generalist robot policy that augments a pretrained vision–language model (PaliGemma backbone) with a continuous-action flow-matching head. Built on top of a large VLM to inherit Internet-scale semantic knowledge, $\pi_0$ is trained on diverse real-robot data and is designed for high-frequency control in dexterous manipulation tasks. The model decodes continuous motor commands via a flow-matching process with 10 integration steps. We use $\pi_0$ as a diffusion-based VLA backbone to evaluate the cross-architecture generalizability of VLA-Pruner. To capture informative interactions and reduce noise, we record the action-to-vision cross-attention scores averaged over the latter half of these integration steps.

\subsubsection{Acceleration Baseline Details}
\label{app:baseline_details}
\begin{figure*}[t]
    \centering
    \subfloat[Tasks on LIBERO benchmark and the SIMPLER environment.\label{fig:libero_simpler}]{%
        \includegraphics[width=0.48\linewidth]{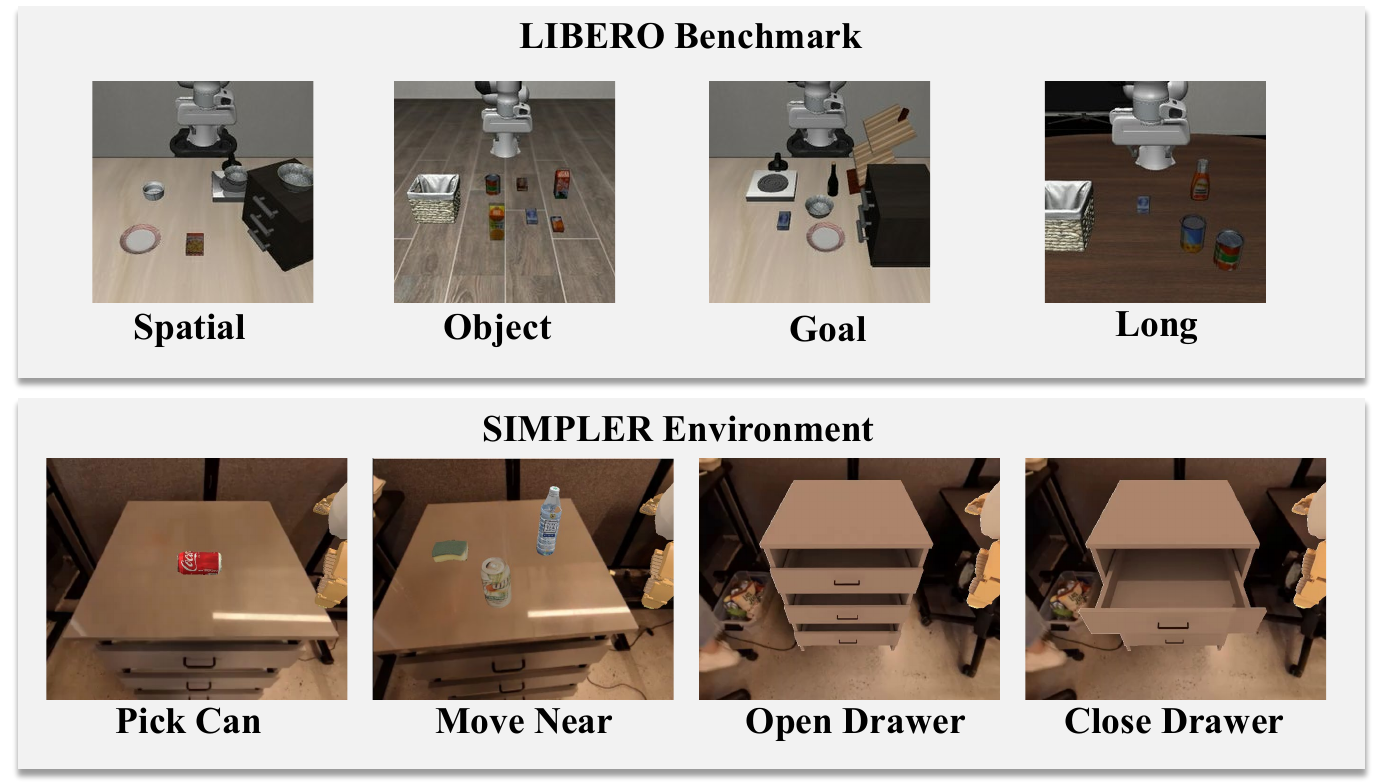}%
    }
    \hfill
    \subfloat[Real-world xArm6 robot setup.\label{fig:robotsetup}]{%
        \includegraphics[width=0.48\linewidth]{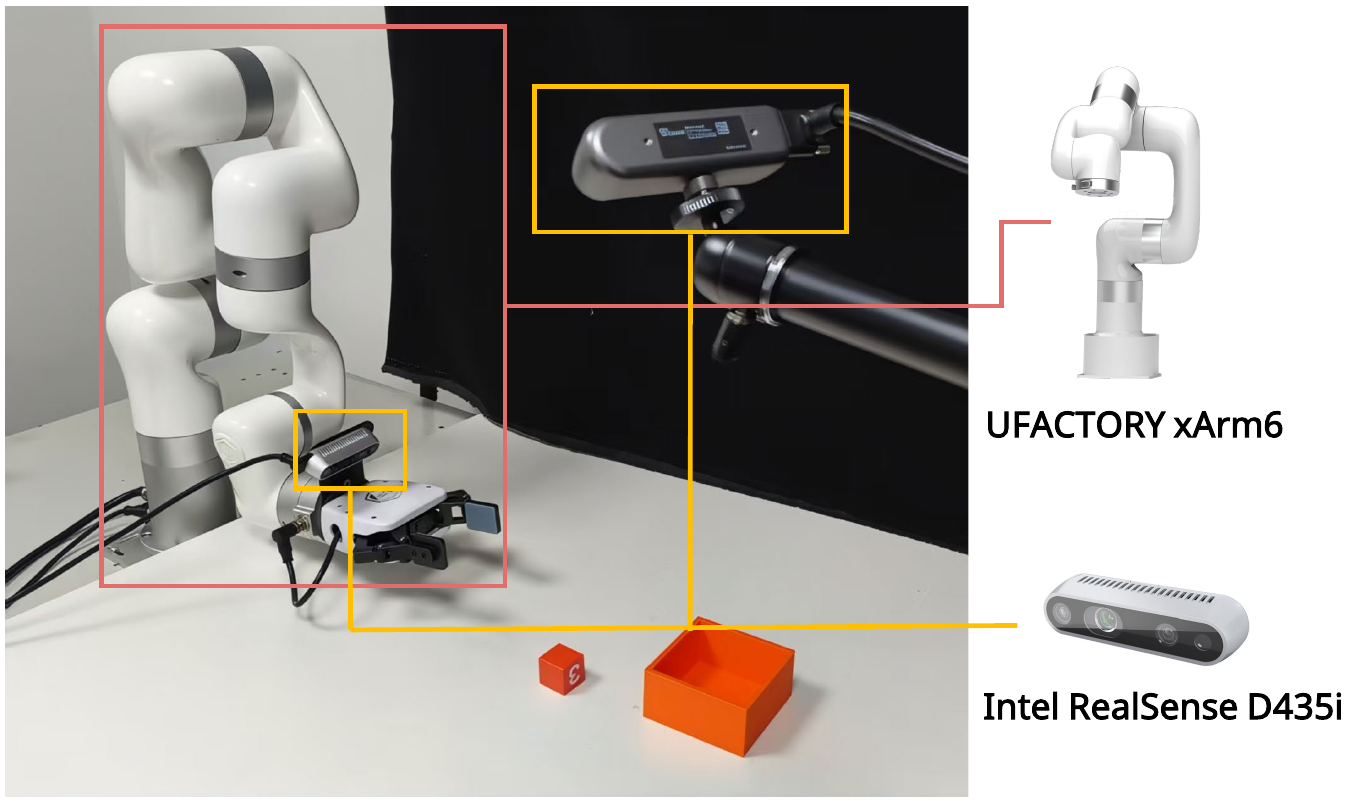}%
    }
    \vspace{-0.4cm}
    \caption{Evaluation environments and real-world robot setup.}
    \label{fig:evaluation_setup}
\end{figure*}
We briefly summarize the baseline acceleration methods considered in our experiments.

\noindent\textbf{FastV.}
FastV~\citep{chen2024fastv}(\texttt{ECCV24}) accelerates large vision–language models by pruning redundant visual tokens based on early-layer prefill attention. Its core observation is that many image patches receive consistently low attention in deeper layers while contributing substantially to computation. FastV defines an attention-based importance score during the prefill stage and drops a fixed fraction of low-score visual tokens after a chosen layer. It is training-free and purely semantic, but does not account for the action-dependent requirements of VLA models.

\noindent\textbf{SparseVLM.}
SparseVLM~\citep{zhang2024sparsevlm}(\texttt{ICML25}) also targets visual token redundancy in VLMs, but uses text-to-vision cross-attention as the importance signal. Tokens that receive low cross-attention from text queries are considered less relevant and pruned. Compared to FastV, SparseVLM focuses directly on aligning visual tokens with language semantics, yet still operates only at the semantic level and ignores the distinct visual needs of action decoding in VLA.

\noindent\textbf{DivPrune.}
DivPrune~\citep{alvar2025divprune}(\texttt{CVPR25}) addresses the tendency of attention-based methods to select similar and  redundant tokens. Instead of ranking tokens by attention scores, it formulates token selection as a Max–Min Diversity Problem and chooses a subset of visual tokens that are maximally dissimilar in feature space, thereby reducing redundancy while maintaining coverage. However, DivPrune is agnostic to the dual-level relevance of visual tokens in VLA prefilling and action decoding, and its diversity objective falls short for VLA tasks where localized details are critical.

\noindent\textbf{VLA-Cache.}  
VLA-Cache~\citep{xu2025vlacache}(\texttt{Neurips25}) is a training-free acceleration method tailored for VLA models. It exploits temporal continuity in robotic manipulation by identifying visually static tokens whose features change little across timesteps, and reuses their computation via a KV-cache. At the same time, it uses text-to-vision cross-attention to keep task-relevant tokens fully computed. This cache-based mechanism reduces repeated computation over static background regions, but its semantic attention score based token selection remains coarse-grained, which overlooks distinct action attention distribution.
To ensure a fair comparison, we set the dynamic token limit to be $1.5 \times k$ ($k$ is the token budget), and use text-to-vision score to exclude $0.5\times k$ of the most relevant tokens. These settings follow the default setup of VLA-Cache codebase.

\noindent\textbf{EfficientVLA.}
EfficientVLA~\citep{yang2025efficientvla}(\texttt{Neurips25}) proposes a structured acceleration framework for diffusion-based VLA models that combines visual token pruning, layer reduction, and diffusion temporal feature caching. Its token-pruning component leverages semantic attention and diversity heuristics inherited from VLM pruning, while additional modules (i.e., layer skipping and diffusion caching) are used to achieve comparable acceleration. In contrast, VLA-Pruner is a single, plug-and-play token pruning module. These approaches are therefore complementary: EfficientVLA and other VLA acceleration methods can be adopted orthogonally with VLA-Pruner, while VLA-Pruner achieves strongest token-prune performance.\\

\begin{figure*}[h]
    \centering
    \subfloat{%
        \includegraphics[width=0.22\textwidth]{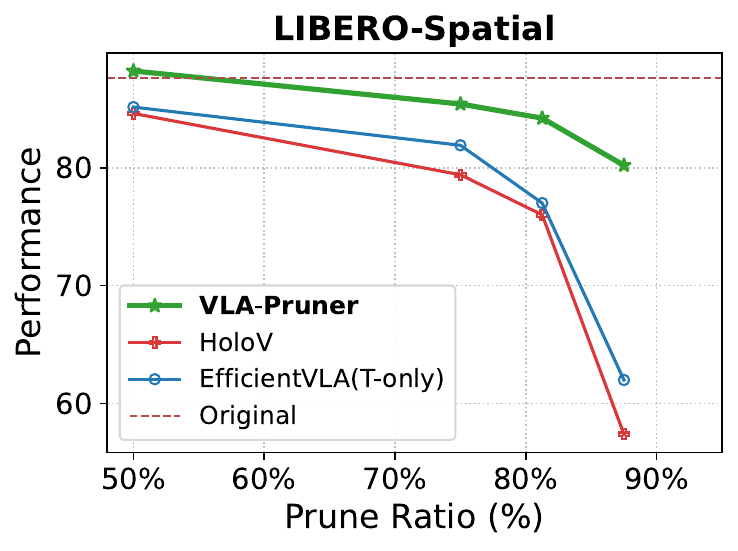}%
    }\hfill
    \subfloat{%
        \includegraphics[width=0.22\textwidth]{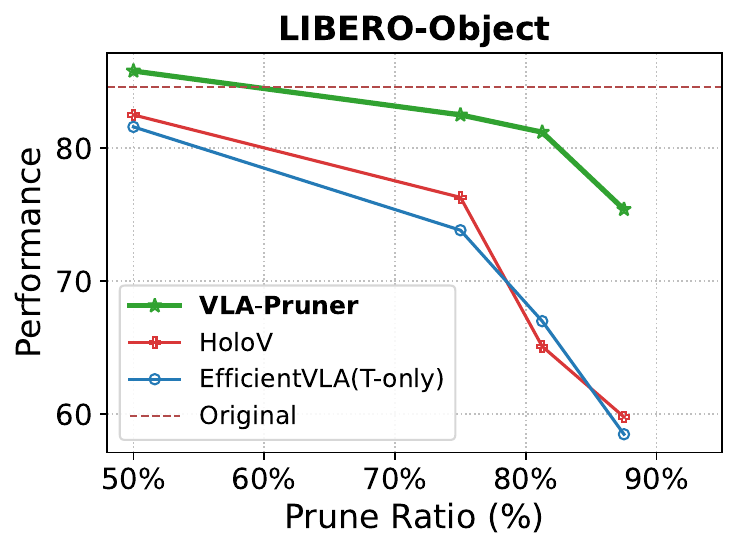}%
    }\hfill
    \subfloat{%
        \includegraphics[width=0.22\textwidth]{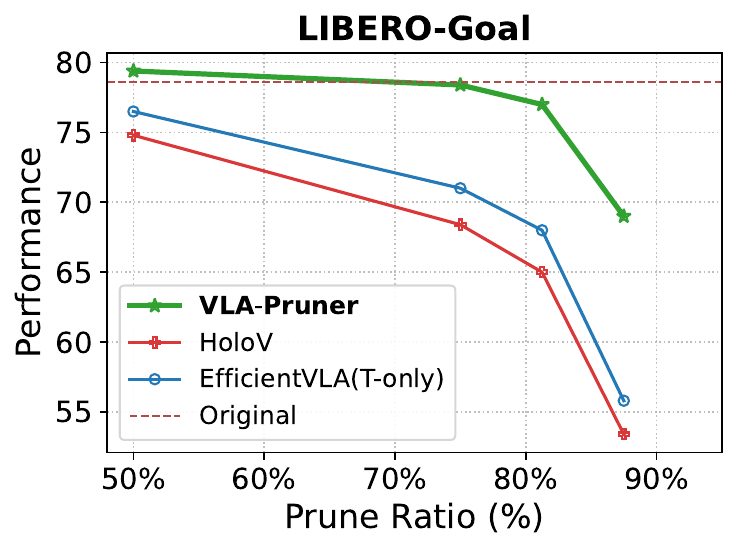}%
    }\hfill
    \subfloat{%
        \includegraphics[width=0.22\textwidth]{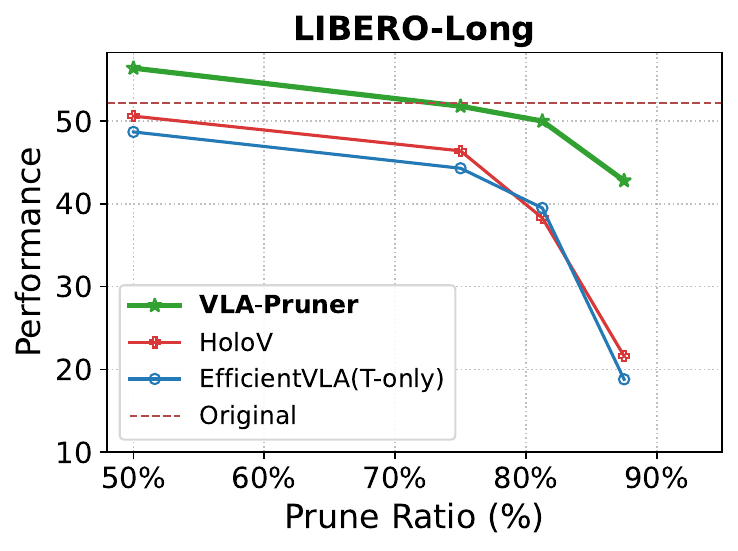}%
    }
    \caption{Performance of VLA-Pruner with more token pruning baseline methods across LIBERO tasks under varying pruning ratios. The horizontal axis represents the pruning/caching ratio of visual tokens, and the vertical axis shows the success rate.}
    \label{fig:appvary}
\end{figure*}
\vspace{-0.4cm}
\begin{table*}[t]
\centering
\caption{Performance comparison with various acceleration methods.}
\setlength{\tabcolsep}{4pt}        
\renewcommand{\arraystretch}{0.85} 
\footnotesize                      
\begin{tabular}{lccccccc}
\toprule
Method & Spatial & Object & Goal & Long & Avg.$\uparrow$ & FLOPs$\downarrow$ & Speedup$\uparrow$ \\
\midrule
VLA-Pruner              & 85.4 & 82.5 & 78.4 & 51.8 & 74.5 & 39.80\% & 1.633$\times$ \\
EfficientVLA (T-only)& 84.3 & 81.6 & 76.5 & 49.8 & 73.1 & 58.12\% & 1.334$\times$ \\
EfficientVLA            & 83.2 & 80.3 & 75.7 & 49.1 & 72.0   & 41.30\% & 1.491$\times$ \\
VTW                     & 74.7 & 76.2 & 72.3 & 48.9 & 68.0 & 55.17\% & 1.325$\times$ \\
FitPrune                & 82.4 & 80.1 & 74.8 & 50.3 & 71.9   & 47.71\% & 1.397$\times$ \\
\bottomrule
\end{tabular}

\label{tab:extended_baselines}
\end{table*}

\subsection{Benchmark Details}
\vspace{-0.1cm}
Fig.~\ref{fig:libero_simpler} shows the evaluation simulation environments and Fig.~\ref{fig:robotsetup} presents our real-world xArm6 robot setup.
\vspace{-0.4cm}
\label{app:benchmark details}
\paragraph{LIBERO.}
LIBERO~\citep{liu2023libero} is a Robosuite-based benchmark for lifelong robot manipulation that groups 130 language-conditioned tasks into four suites: \textit{Spatial}, \textit{Object}, \textit{Goal}, and \textit{Long}. LIBERO-Spatial varies only the spatial configuration of otherwise identical objects (e.g., pick up the black bowl on the stove and place it on the plate). LIBERO-Object varies the manipulated object while keeping the scene layout and goal template fixed (e.g., pick up the bbq sauce and place it in the basket). LIBERO-Goal keeps both the objects and spatial layout fixed but changes the goal description (e.g., open the middle drawer of the cabinet). LIBERO-Long corresponds to the long-horizon subset (10 tasks) of LIBERO-100, where each episode consists of multi-stage instructions such as opening a drawer, inserting an object, and then rearranging other items(e.g., put both the alphabet soup and the cream cheese box in the basket). 
\vspace{-0.3cm}
\paragraph{SIMPLER.}
SIMPLER~\citep{li24simpler} is a simulation suite built to closely track the real Google robot environment while allowing controlled appearance variation. It provides two evaluation settings. The \textit{Visual Matching} (VM) setting keeps backgrounds, lighting, distractors, and camera poses close to the real setup so that rendered observations visually resemble real robot scenes and better predict real-world performance. The \textit{Variant Aggregation} (VA) setting starts from the same base scenes but procedurally perturbs nuisance factors such as background textures, illumination, number and placement of distractor objects, table material, and camera viewpoints, creating a broad distribution shift for testing robustness. For the Google robot configuration, SIMPLER defines four canonical manipulation tasks: “pick coke can”,“pick object”, “move near”, “open/close Drawer”, and “place in closed drawer”. In our experiments, we follow the official simulator settings and success metrics, and select three representative tasks—\textit{Pick Coke Can}, \textit{Move Near}, and \textit{Open/Close Drawer}—under VM or VA setting, where base model attains non-trivial performance.
\vspace{-0.3cm}
\paragraph{Real robot.}
As shown in Figure~\ref{fig:robotsetup}, our real-world system is built on a 6-DoF xArm6 manipulator equipped with two Intel RealSense D435i RGB cameras: one mounted on the wrist for close-range observations and the other fixed for capturing global scene context. The robot is controlled through the official xArmAPI. We collect real-world demonstrations following the LeRobot2.1 data specification. Each episode consists of synchronized observations recorded at 30Hz, including two RGB image streams. The proprioceptive state is represented by the robot's absolute joint angles, and actions are stored using the same absolute joint angles to ensure consistent replay and supervision. All demonstrations are gathered using this unified multi-view setup, which provides both global scene context and fine-grained manipulation details that are essential for learning robust visuomotor policies.

\begin{table*}[t]
\centering
\footnotesize
\caption{\textbf{Performance of VLA-Pruner on $\pi_0$ across LIBERO at different vision-token retention ratios.} We report success rates (\%) on the four suites, relative accuracy, latency (ms), speedup ($\times$) and FLOP ratio (\%).}
\vspace{-0.3cm}
\label{tab:pi0_libero_full}
\setlength{\tabcolsep}{4pt} 
\renewcommand{\arraystretch}{0.82} %
\begin{tabular}{lccccccccc}
\toprule
\multirow{2}{*}{Method} & \multicolumn{5}{c}{LIBERO Success (\%)} & \multirow{2}{*}{Acc.(\%)$\uparrow$} & \multirow{2}{*}{Latency(ms)$\downarrow$} & \multirow{2}{*}{Speedup ($\times$)$\uparrow$} & \multirow{2}{*}{FLOP ratio(\%)$\downarrow$} \\
\cmidrule(lr){2-6}
& Spatial & Object & Goal & Long & Avg. & & & & \\
\hline
\rowcolor{gray!13}
\multicolumn{10}{c}{\textit{Upper Bound (100\%)}} \\
\hline
Vanilla & 96.9 & 98.3 & 96.1 & 84.8 & 94.03 & 100.0 & 104.53 & 1.000 & 100 \\
\hline
\rowcolor{gray!12}
\multicolumn{10}{c}{\textit{Retain 50\% Tokens}} \\
\hline
FastV & 95.26 & 95.46 & 92.58 & 81.34 & 91.16 & 91.16 & 66.44 & 1.573 & 60.7 \\
SparseVLM & 94.70 & 93.96 & 91.50 & 79.48 & 89.91 & 89.91 & 70.39 & 1.485 & 61.5 \\
DivPrune & 91.30 & 92.14 & 88.25 & 77.19 & 87.22 & 87.22 & 69.17 & 1.511 & 58.9 \\
VLA-Cache & 96.47 & 95.99 & 96.59 & 84.65 & 93.43 & 93.43 & 78.30 & 1.335 & 72.3 \\
\textbf{VLA-Pruner} & 97.01 & 98.61 & 97.57 & 87.12 & 95.07 & \textbf{101.10} & 69.45 & 1.505 & 60.9 \\
\hline
\rowcolor{gray!12}
\multicolumn{10}{c}{\textit{Retain 25\% Tokens}} \\
\hline
FastV & 88.28 & 78.81 & 86.38 & 73.61 & 81.77 & 86.97 & 57.68 & 1.812 & 39.9 \\
SparseVLM & 89.63 & 84.78 & 81.80 & 72.67 & 82.22 & 86.76 & 61.09 & 1.711 & 41.3 \\
DivPrune & 84.41 & 71.60 & 80.61 & 66.60 & 75.80 & 80.76 & 59.05 & 1.770 & 39.3 \\
VLA-Cache & 85.60 & 85.32 & 85.94 & 73.97 & 82.71 & 87.86 & 64.81 & 1.613 & 51.4 \\
\textbf{VLA-Pruner} & 95.28 & 96.25 & 95.71 & 83.01 & 92.56 & \textbf{98.44} & 60.38 & 1.731 & 40.1 \\
\hline
\rowcolor{gray!12}
\multicolumn{10}{c}{\textit{Retain 12.5\% Tokens}} \\
\hline
FastV & 68.31 & 59.15 & 57.97 & 28.08 & 53.38 & 53.37 & 51.72 & 2.021 & 30.1 \\
SparseVLM & 72.04 & 64.98 & 63.55 & 30.99 & 57.89 & 57.89 & 53.79 & 1.943 & 31.3 \\
DivPrune & 63.41 & 63.78 & 62.79 & 29.61 & 54.90 & 54.90 & 52.64 & 1.986 & 29.8 \\
VLA-Cache & 59.85 & 58.69 & 64.71 & 27.09 & 52.58 & 55.92 & 59.02 & 1.771 & 39.4 \\
\textbf{VLA-Pruner} & 89.64 & 90.55 & 86.60 & 65.56 & 83.08 & \textbf{88.35} & 53.49 & 1.954 & 30.4 \\
\bottomrule
\end{tabular}
\end{table*}

\begin{table*}[h]
    \centering
    \caption{\textbf{Memory and Runtime Analysis of Acceleration Methods on OpenVLA.} Detailed comparison of maximum GPU memory consumption and CUDA runtime across different vision-token retention rates.}
    \vspace{-0.3cm}
    \label{tab:appendix_memory_time}
    \renewcommand{\arraystretch}{0.85}
    \resizebox{0.96\linewidth}{!}{
        \setlength{\tabcolsep}{6pt}
        \begin{tabular}{l|cc|cc|cc}
            \hline
            \multirow{2}{*}{\textbf{Method}} & \multicolumn{2}{c|}{\textbf{50\% Retained}} & \multicolumn{2}{c|}{\textbf{25\% Retained}} & \multicolumn{2}{c}{\textbf{12.5\% Retained}} \\
            \cline{2-7}
             & Max Memory (GB) $\downarrow$ & CUDA Time (ms) $\downarrow$ & Max Memory (GB) $\downarrow$ & CUDA Time (ms) $\downarrow$ & Max Memory (GB) $\downarrow$ & CUDA Time (ms) $\downarrow$ \\
            \hline
            Vanilla & \multicolumn{6}{c}{Origin Memory: 14.75 GB / Origin CUDA Latency: 51.43 ms}\\
            \hline
            FastV & 14.513 & 29.022 & 14.327 & 25.58 & 14.311 & 24.33 \\
            SparseVLM & 14.544 & 29.594 & 14.351 & 25.71 & 14.329 & 23.64 \\
            DivPrune & 14.438 & 28.052 & 14.303 & 24.98 & 14.281 & 22.91 \\
            VLA-Cache & 14.592 & 36.33 & 14.523 & 35.13 & 14.484 & 34.57 \\
            \textbf{VLA-Pruner} & 14.527 & 29.192 & 14.348 & 25.58 & 14.322 & 23.35 \\
            \hline
        \end{tabular}
        \vspace{-0.8cm}
    }
\end{table*}

\subsection{Results of More Baselines}
\label{app:more_baselines}
\noindent\textbf{Varying pruning ratio.}
Besides the main training-free baselines in the paper (FastV~\citep{chen2024fastv}, SparseVLM~\citep{zhang2024sparsevlm}, DivPrune~\citep{alvar2025divprune}, VLA-Cache~\citep{xu2025vlacache}), we evaluate two recent visual token pruning methods with controllable pruning ratios: token-pruning module of EfficientVLA~\citep{yang2025efficientvla}, denoted as \textit{EfficientVLA(T-only)} and HoloV(\texttt{Neurips25})~\citep{holov}. We evaluate them on OpenVLA backbone across four LIBERO suites. For each method, we report the average success rates under varying retained token ratios in Fig.~\ref{fig:appvary}. VLA-Pruner achieves consistent superiority across all suites. These results further support that semantic-only token pruning is ill-suited for VLA, and that our dual-level strategy better matches the requirements of VLA robotic control.

\noindent\textbf{Comparison under comparable performance.}
We compare VLA-Pruner against structured and calibration-based baselines, focusing on the trade-off between VLA performance and computational cost. We consider the full EfficientVLA framework~\citep{yang2025efficientvla}, which combines token pruning and layer reduction, as well as the calibration-based methods FitPrune~\citep{ye2025fitprune} and VTW~\citep{lin2025VTW}.
For EfficientVLA, we set the visual-token retention at $50\%$ and reduce 8 layers to ensure comparable performance while maximizing acceleration. We report VLA-Pruner at $25\%$ token retention, which achieves the better performance while requiring fewer FLOPs and lower latency. This suggests that dual-level VLA token pruning can achieve a better efficiency–performance trade-off than EfficientVLA, which is designed primarily for diffusion head policy and overlook the intrinsic characteristics of VLA inference for token pruning, thereby showing limited performance. For FitPrune and VTW, we follow their official calibration protocols and perform calibration on 100 LIBERO tasks, which corresponds to a substantially larger calibration fraction than that typically used in VLM benchmarks. Both methods achieve lower success rates than VLA-Pruner at $25\%$ token retention and consume more FLOPs (Tab.~\ref{tab:extended_baselines}). VTW typically chooses a late withdrawal layer and degrades VLA performance, while FitPrune, though effective on standard VLM benchmarks, shows limited favorable trade-off in the VLA setting since it stills rely on semantic attention scores.
\subsection{Results of $\pi_0$ on LIBERO}
\label{app:pi0_libero}
\noindent\textbf{Setup.} We evaluate VLA-Pruner on the $\pi_0$ model across LIBERO benchmark at various token retention ratios (50\%, 25\%, and 12.5\%). For each configuration, we measure success rates, relative accuracy, latency, speedup, and FLOP ratios. We set the number of flow matching integration steps as 10, action chunk as 50 and execute 5 actions per step.

\noindent\textbf{Main Results.}
VLA-Pruner consistently achieves high performance across all token retention settings. Notably, at each tested token retention ratio, our method maintains superior performance relative to other baselines, especially as pruning ratio increases. In particular, for $\pi_0$, our method demonstrates impressive efficiency, maintaining comparable success rates even at 12.5\% token retention. This enables a significant acceleration in computation while preserving performance, a key advantage over traditional pruning methods.
While VLA-Pruner's redundancy filtering slightly affects acceleration performance on $\pi_0$ with high-frequency, the impact remains minimal. Even at a 25\% retention ratio, VLA-Pruner surpasses other token pruning methods at 50\% retention ratio, demonstrating that our method can precisely preserve essential information. The results in Table~\ref{tab:pi0_libero_full} clearly illustrate how VLA-Pruner maintains robust task success across pruning ratios, highlighting its cross-architecture generalizability to provide an superior trade-off between computational efficiency and model accuracy for diffusion-based model.

\subsection{More results}
\label{app:more_results}

\subsubsection{Detailed Efficiency Results}
\label{app:more_results:memory}
We report additional efficiency results, including peak GPU memory consumption (GB) and CUDA runtime (ms), across different vision-token retention ratios. As shown in Table~\ref{tab:appendix_memory_time}, VLA-Pruner exhibits comparable efficiency to other pruning methods.

\subsubsection{More Analysis}
\label{app:more_results:ablations}
\begin{figure}[t]
    \centering
    \begin{minipage}[t]{0.44\linewidth}
        \centering
        \includegraphics[width=\linewidth]{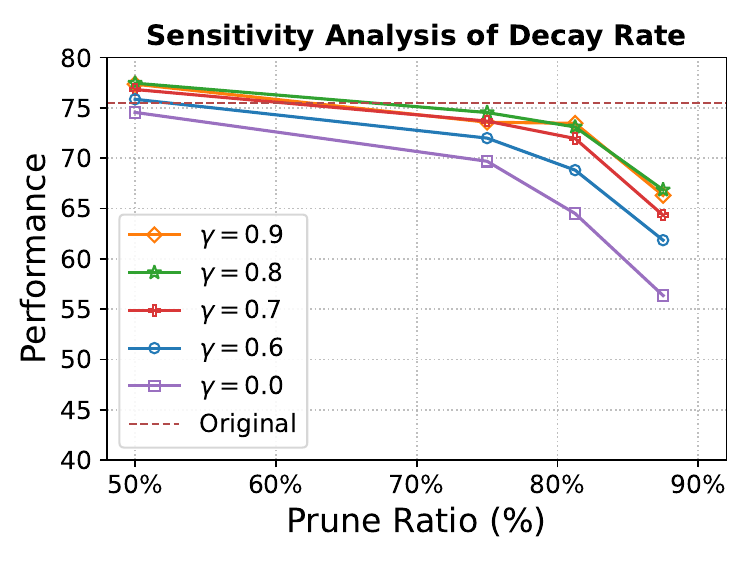} 
        \vspace{-0.8cm}
        \caption{\footnotesize Ablation study for decay rate.}
        \label{fig:ablation_decay}
    \end{minipage}
    \hfill 
    \begin{minipage}[t]{0.52\linewidth}
        \centering
        \includegraphics[width=\linewidth]{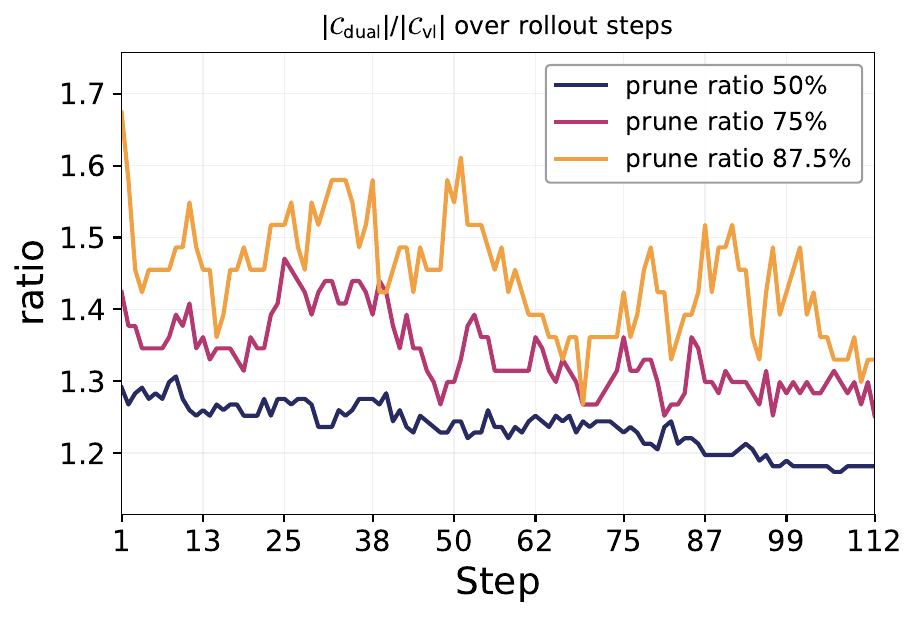} 
        \vspace{-0.8cm}
        \caption{\footnotesize $|\mathcal{C}_{\text{dual}}|/|\mathcal{C}_{\text{vl}}|$ ratio trends at different pruning ratios on robot task.} 
        \label{fig:dual} 
    \end{minipage}
    \vspace{-0.3cm} 
\end{figure}

\noindent\textbf{Decay rate $\gamma$.}  
We analyze how different decay rates $\gamma \in \{0.0, 0.6, 0.7, 0.8, 0.9\}$ influence VLA-Pruner and report the results for average success rates under varying prune ratio in Fig.~\ref{fig:ablation_decay}.\\
\noindent\textbf{Pruning layer $K$.}  
We compare results of different pruning layer values $K \in \{2, 3, 4, 5\}$ in Table~\ref{tab:ablation_layers} (retain 25\% tokens).\\ 
\noindent\textbf{VLA-Pruner details.}  
We show $|\mathcal{C}_{\text{dual}}|/|\mathcal{C}_{\text{vl}}|$ ratio trends at different pruning ratios for a robotic task in Fig.~\ref{fig:dual}. 
\begin{table}[t]
    \centering
    \setlength{\tabcolsep}{4pt}
    \renewcommand{\arraystretch}{0.9}
    \resizebox{0.7\linewidth}{!}{
        \begin{tabular}{l|c|cccc|c}
            \toprule
            Method & TFLOP (ratio \%) & Spatial & Object & Goal & Long & Avg \\ 
            \midrule
            K=2         & 37.90 & 69.7 & 74.1 & 55.3 & 34.8 & 58.48 \\
            K=3 (Ours)  & 39.77 & 85.4 & 82.5 & 78.4 & 51.8 & 74.53 \\
            K=4         & 41.98 & 85.5 & 82.3 & 78.1 & 52.0 & 74.48 \\
            K=5         & 43.08 & 83.9 & 81.7 & 77.9 & 51.4 & 73.73 \\
            \bottomrule
        \end{tabular}
    }
    \caption{Study on applying pruning at different layers.}
    \label{tab:ablation_layers}
    \vspace{-0.4cm}
\end{table}

\clearpage
\subsection{More Visualization}
\begin{figure*}[h]
    \centering
    \includegraphics[width=0.7\linewidth]{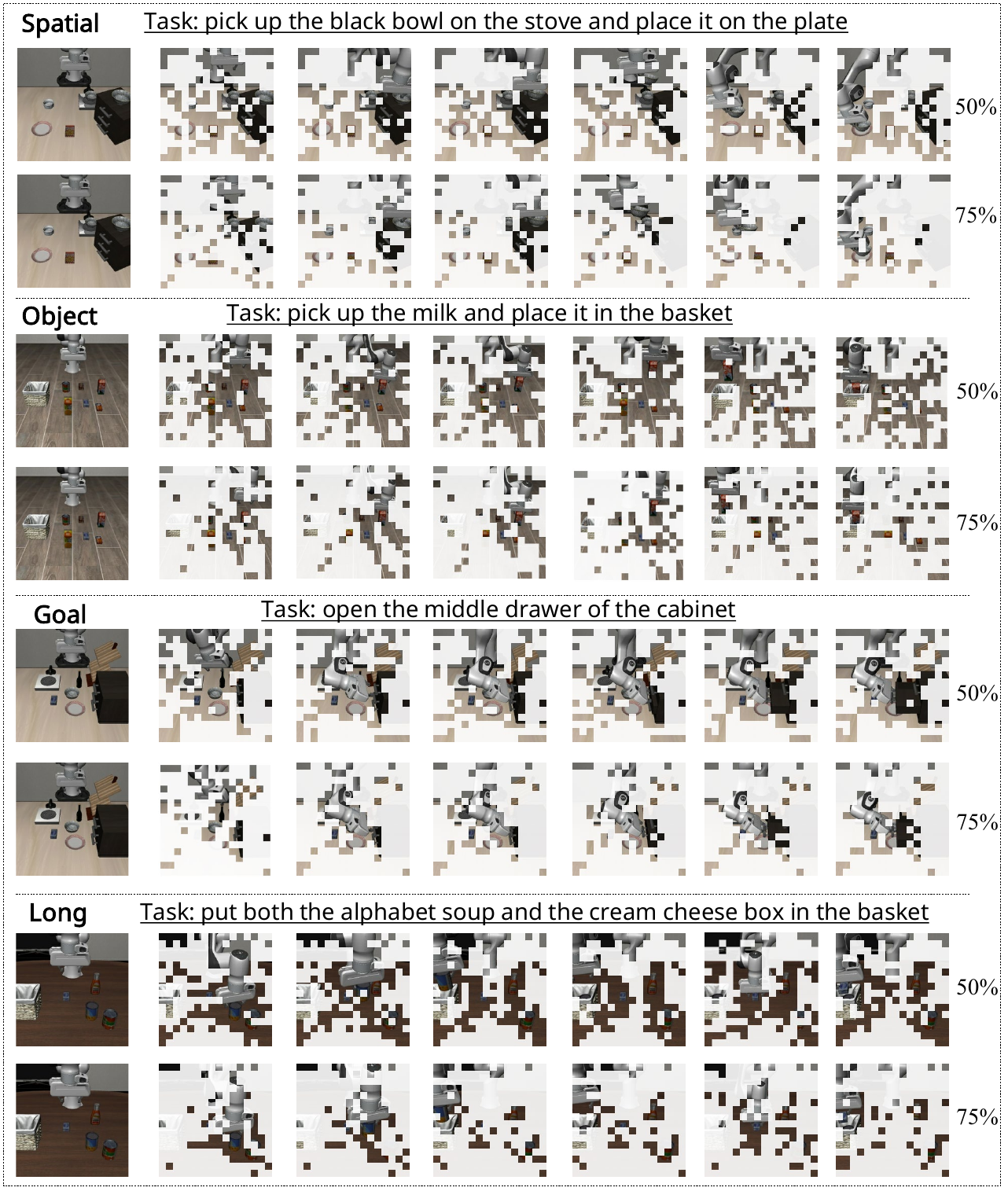}
    \caption{Visualization of VLA-Pruner token pruning results across LIBERO tasks under different ratios}
    \label{fig:more}
\end{figure*}

We provide detailed visualization of VLA-Pruner token prune results across LIBERO tasks (50\%, 75\% pruning ratios) in Fig.~\ref{fig:more}. VLA-Pruner precisely maintains both local details for action execution and global information for semantic understanding and task planning.

\section{Limitations and Future work.}
\label{app:limitations}
Our main contribution is to point out the dual-system nature of VLA inference and to propose a dual-level importance criterion that jointly accounts for semantic and action-level relevance. However, our current estimation of action-level importance relies on a heuristic temporal smoothing scheme with a fixed window and decay rate. While this works well for most tabletop settings, its benefits may diminish in highly dynamic scenarios (e.g., egocentric wrist-camera views or conveyor-belt environments), where rapid viewpoint shifts and object motion can violate the short-term temporal continuity assumption and blur sharp changes in action-to-vision attention. A promising direction for future work is to replace the fixed temporal smoothing with an adaptive prediction module that adjusts its effective window based on motion cues or feature changes (e.g., magnitude of visual dynamics or action-conditioned attention variance), or to learn a lightweight temporal attention network that directly predicts per-token importance from recent history. Such adaptive mechanisms may further improve pruning quality in dynamic environments while retaining the advantages of VLA-Pruner.


\end{document}